%% file: acl_latex.tex
\pdfoutput=1

\documentclass[11pt]{article}

\usepackage[preprint]{acl}

\usepackage{times}
\usepackage{latexsym}

\usepackage[T1]{fontenc}

\usepackage[utf8]{inputenc}
\usepackage{graphicx}
\usepackage{tikz}
\usepackage{forest}
\usetikzlibrary{trees,positioning,shapes,shadows,arrows.meta}
\usepackage{amsmath,amsfonts,amssymb}
\usepackage{todonotes}
\usepackage{subfig}
\usepackage{arydshln}
\usepackage{enumitem}

\usepackage{tikz}
\usepackage{tcolorbox}

\usepackage{microtype}

\usepackage{inconsolata}

\usepackage{graphicx}
\usepackage{enumitem}

\usepackage{multirow}
\usepackage{tabularx}
\usepackage{longtable}
\usepackage{array}
\usepackage{lineno}

\title{How to Upscale Neural Networks with Scaling Law? \\A Survey and Practical Guidelines}

\author{
   Ayan Sengupta$^*$\hspace{10pt}
   Yash Goel$^*$\hspace{10pt} 
   Tanmoy Chakraborty
   \\
   Indian Institute of Technology Delhi, India
   \\
   \texttt{\{ayan.sengupta, ee1210984, tanchak\}@ee.iitd.ac.in}
 }

\begin{document}

\addtocontents{toc}{\protect\setcounter{tocdepth}{-1}}

\maketitle

\def\thefootnote{*}
\textsuperscript{*}\footnotetext[0]{Equal contribution}
\def\thefootnote{\arabic{footnote}}

\begin{abstract}
Neural scaling laws have revolutionized the design and optimization of large-scale AI models by revealing predictable relationships between model size, dataset volume, and computational resources. Early research established power-law relationships in model performance, leading to compute-optimal scaling strategies. However, recent studies highlighted their limitations across architectures, modalities, and deployment contexts. Sparse models, mixture-of-experts, retrieval-augmented learning, and multimodal models often deviate from traditional scaling patterns. Moreover, scaling behaviors vary across domains such as vision, reinforcement learning, and fine-tuning, underscoring the need for more nuanced approaches. In this survey, we synthesize insights from over 50 studies, examining the theoretical foundations, empirical findings, and practical implications of scaling laws. We also explore key challenges, including data efficiency, inference scaling, and architecture-specific constraints, advocating for adaptive scaling strategies tailored to real-world applications. We suggest that while scaling laws provide a useful guide, they do not always generalize across all architectures and training strategies.

\end{abstract}

\section{Introduction}

\begin{figure}[!t]
    \centering
    \includegraphics[width=\linewidth]{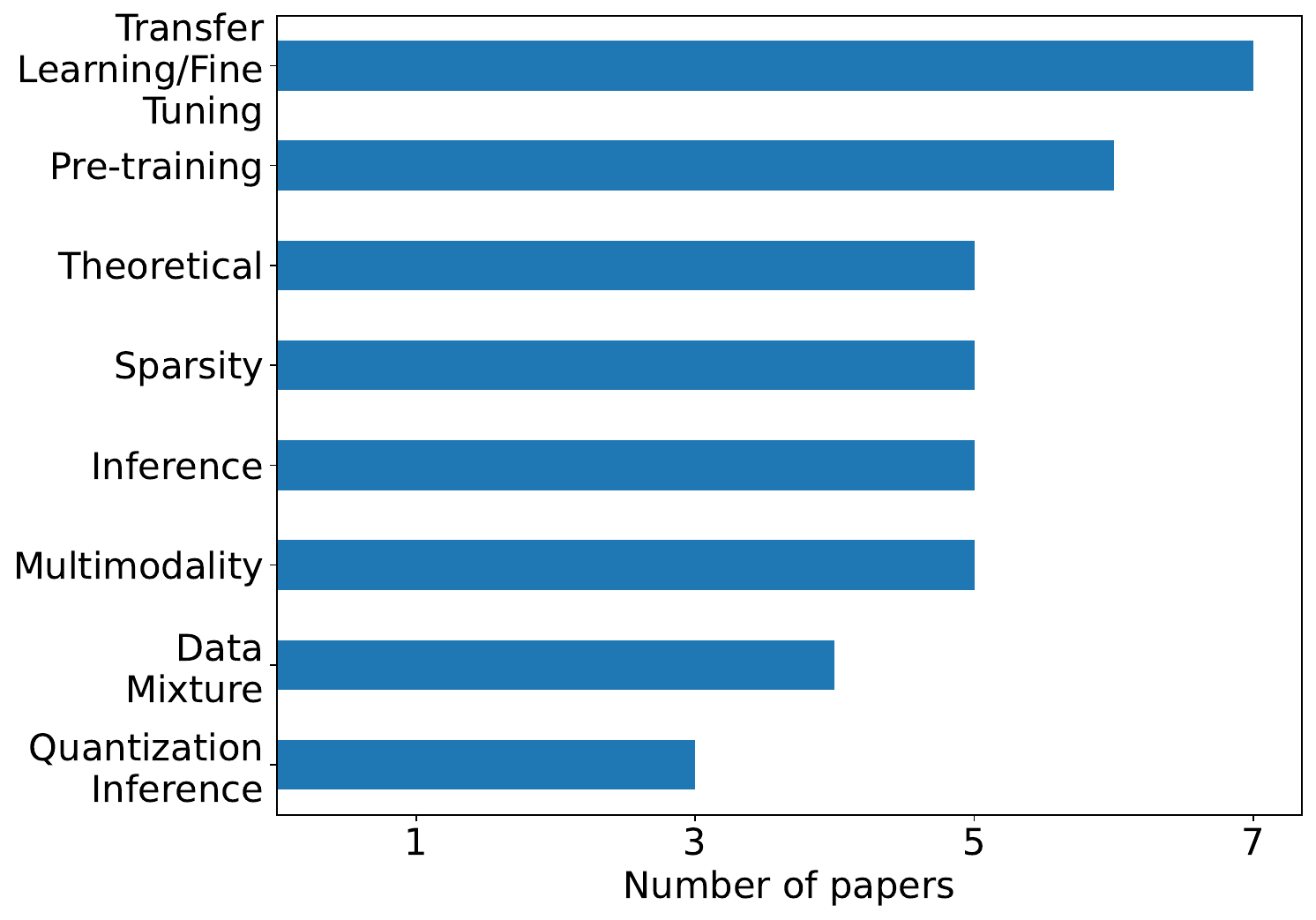}
    \caption{Papers surveyed under different categories. The detailed paper list is provided in Table~\ref{tab:database1} of Appendix~\ref{appx:details}.}
    \label{fig:paper_count}
    \vspace{-5mm}
\end{figure}

\text{Scaling laws} have become a fundamental aspect of modern AI development, especially for large language models (LLMs). In recent years, researchers have identified consistent relationships between model size, dataset volume, and computational resources, demonstrating that increasing these factors leads to systematic improvements in performance. These empirical patterns have been formalized into mathematical principles, known as \textit{scaling laws}, which provide a framework for understanding how the capabilities of neural networks evolve as they grow. Mastering these laws is crucial for building more powerful AI models, optimizing efficiency, reducing costs, and improving generalization.

\begin{figure*}[t]
    \centering
    \includegraphics[width=\textwidth]{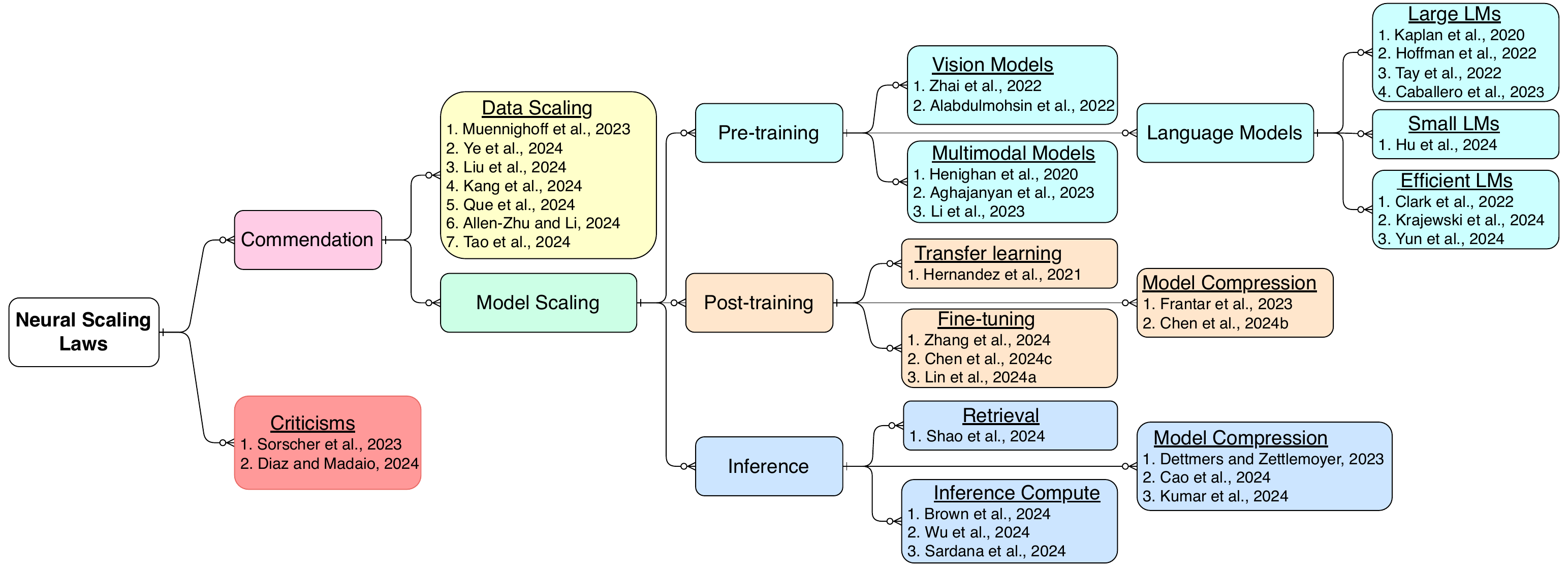}
\caption{A taxonomy of neural scaling laws.}
\label{fig:lit_surv}
\end{figure*}

The study of neural scaling laws gained prominence with the foundational work of~\citet{kaplan_scaling_2020}, who demonstrated that model performance follows a power-law relationship with respect to size, data, and compute. Their findings suggested that larger language models (LMs) achieve lower loss when trained on sufficiently large datasets with increased computational resources. Later, \citet{hoffmann_training_2022} refined these ideas, introducing the notion of compute-optimal scaling, which revealed that training a moderate-sized model on a larger dataset is often more effective than scaling model size alone. However, recent studies~\cite{muennighoff_scaling_2023,caballero_broken_2023,krajewski_scaling_2024} have challenged the universality of these laws, highlighting cases where sparse models, mixture-of-experts architectures, and retrieval-augmented methods introduce deviations from traditional scaling patterns. These findings suggested that while scaling laws provide a useful guide, they do not always generalize across all architectures and training strategies.

Despite the growing importance of scaling laws, existing research remains fragmented, with limited synthesis of theoretical foundations, empirical findings, and practical implications. Given the rapid evolution of this field, there is a need for a structured analysis that consolidates key insights, identifies limitations, and outlines future research directions. While theoretical studies have established the mathematical principles governing scaling, their real-world applications, such as efficient model training, optimized resource allocation, and improved inference strategies, are less explored. To address this gap, we reviewed over 50 research articles (Figure~\ref{fig:paper_count} highlights papers on scaling laws on different topics) to comprehensively analyze scaling laws, examining their validity across different domains and architectures.

While prior surveys have made valuable contributions to understanding scaling laws, they have primarily focused on specific aspects of the scaling phenomenon (See Table \ref{tab:survey_differences}).~\citet{choshen2024hitchhikersguidescalinglaw} emphasized statistical best practices for estimating and interpreting scaling laws using training data, while \citet{li_misfitting_2024} emphasized on methodological inconsistencies and reproduction crisis in existing scaling laws. Our survey distinguishes itself by offering comprehensive coverage of architectural considerations, data scaling implications, and inference scaling -- areas that previous surveys either overlooked or addressed only partially. 

\input{tables/survey_differences}
\label{tab:survey_differences}


\begin{figure*}[!t]
    \centering
    \subfloat[Architecture-wise statistics]{\includegraphics[width=0.33\linewidth]{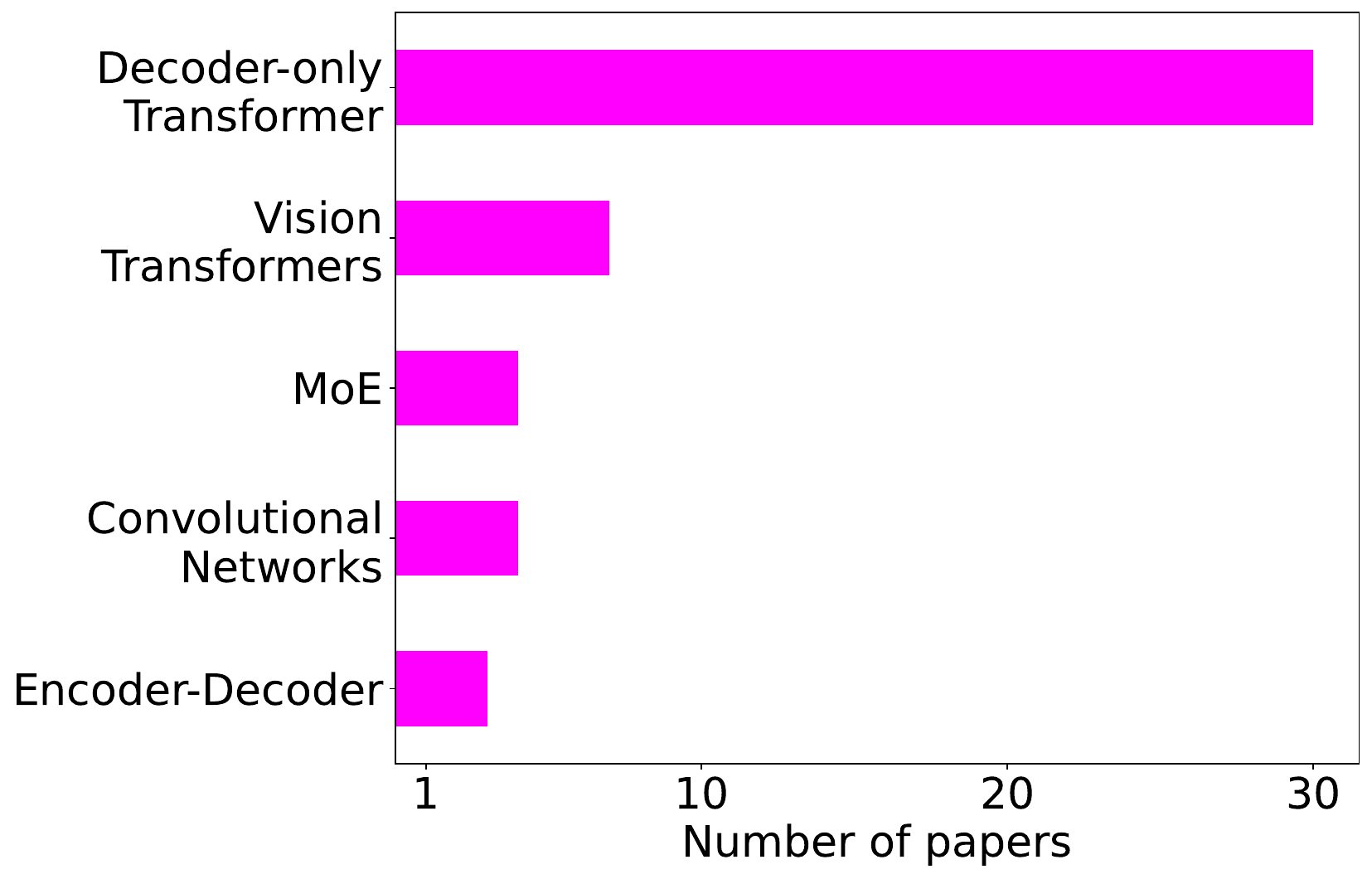}\label{fig:architecture}}
    \subfloat[Variable-wise statistics]{\includegraphics[width=0.31\linewidth]{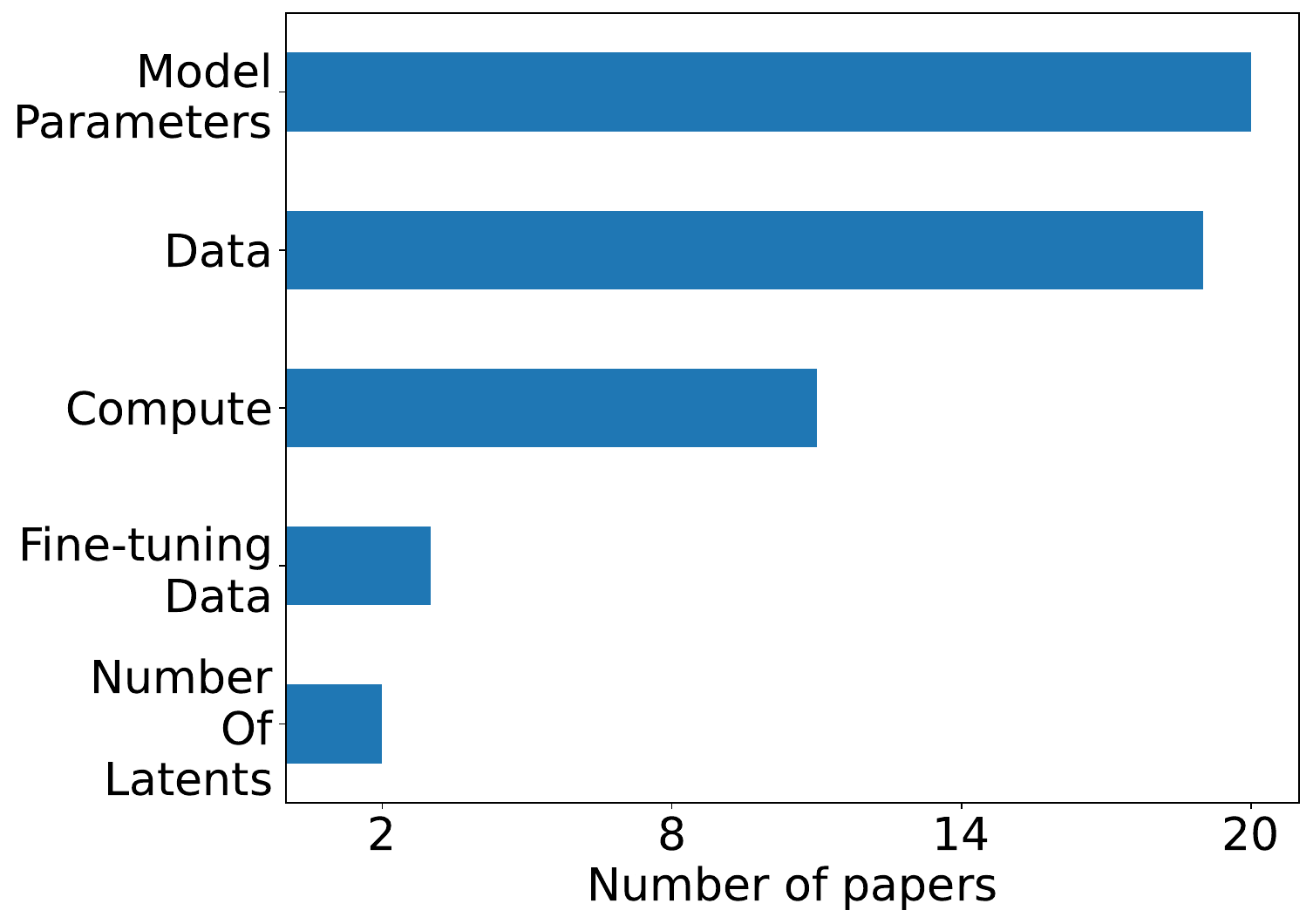}\label{fig:scaling_variable}}
    \subfloat[Task-wise statistics]{\includegraphics[width=0.32\linewidth]{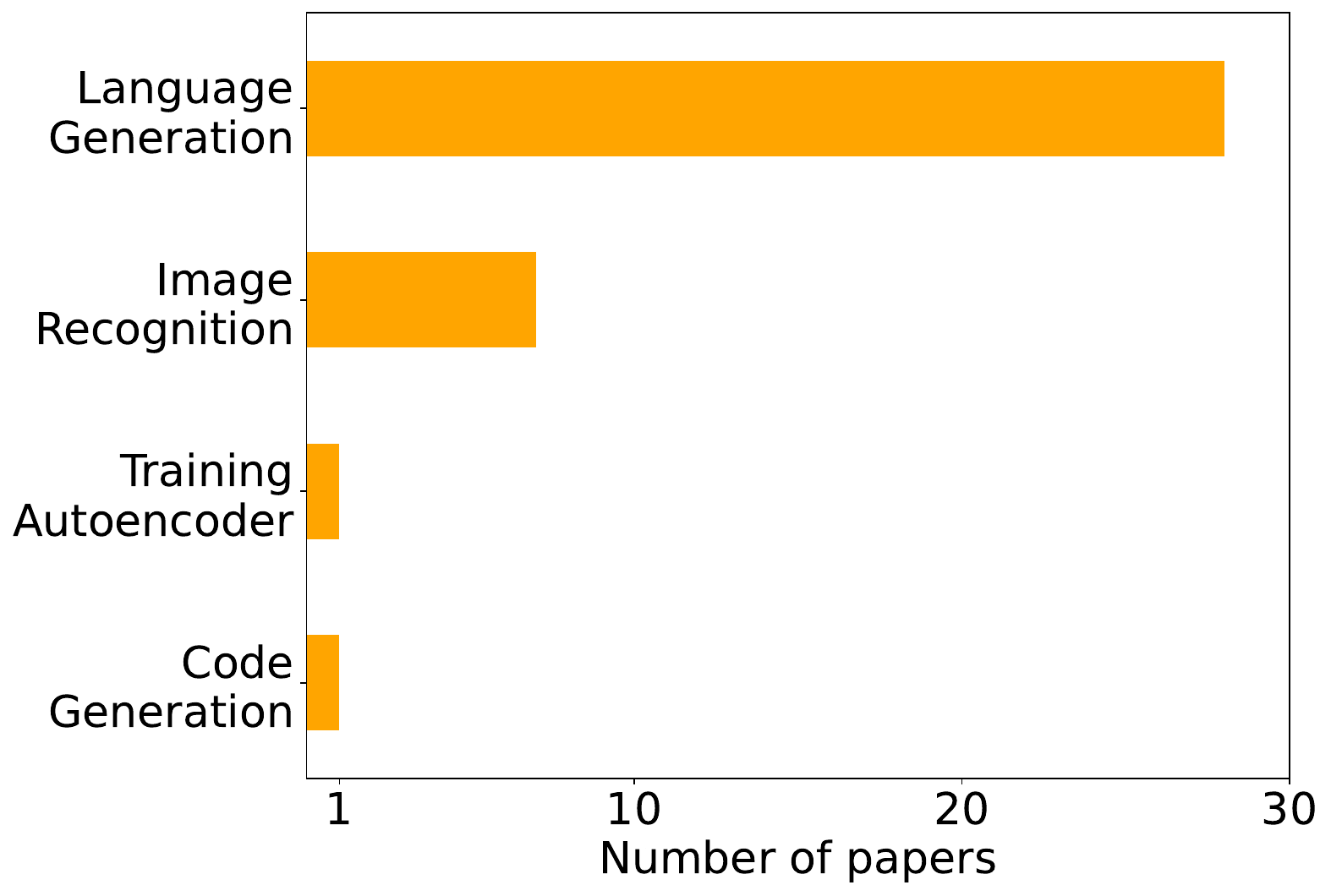}\label{fig:tasks}}
    \caption{Number of paper studied in this survey paper for different model architectures (a), scaling variables (b) and scaling tasks (c). The detailed paper list is provided in Table~\ref{tab:database1} of Appendix~\ref{appx:details}. }
\end{figure*}

\section{Taxonomy of neural scaling laws}
\label{sec:lit_surv}
Understanding the scaling laws of neural models is crucial for optimizing performance across different domains. We predominantly explore the scaling principles for language models, extending to other modalities such as vision and multimodal learning. We also examine scaling behaviors in domain adaptation, inference, efficient model architectures, and data utilization. We highlight the taxonomy tree of scaling laws research in Figure~\ref{fig:lit_surv}. As highlighted in Figure~\ref{fig:paper_count}, neural scaling laws have been proposed predominantly for pre-training and fine-tuning scaling of large neural models. Among the models studied, as highlighted in Figure~\ref{fig:architecture}, decoder-only Transformers dominate the subject, followed by vision transformers (ViT) and Mixture-of-Experts (MoE). 

The most common neural scaling laws take the form of power laws (Equation~\ref{eq:basic_scaling}), where the model's loss ($L$) or performance metric assumes to follow a predictable relationship with different scaling variables, 
\begin{equation}
\small
L(P_{i....n}) = \sum_{i=1}^{n} \alpha_i \cdot P_i^{-\beta_i}
\label{eq:basic_scaling}
\end{equation}
with appropriate scaling parameters $\beta_i$ and fitting parameters $\alpha_i$ for different scaling parameter $P_i$. Figure~\ref{fig:scaling_variable} highlights that the number of model parameters and data size are the most common used scaling factors. The exact forms of all the scaling laws are highlighted in Table~\ref{tab:database2} of Appendix~\ref{appx:details}. Among all the tasks, Figure~\ref{fig:tasks} suggests that language generation is the most common task used for developing these scaling laws, where the training cross-entropy loss is widely used to fit the laws. Based on the values obtained empirically, the scaling laws are fitted with non-linear optimization, most commonly by running algorithms like least square and BFGS (Broyden-Fletcher-Goldfarb-Shanno). Statistical methods like goodness-of-fit metrics are used to validate the correctness of the fitted curves. We elaborate on the evaluation of neural scaling laws in Appendix~\ref{appx:validation}. In the following sections, we review the existing literature on neural scaling across various domains.

\noindent \textbf{Model scaling} includes both parameter and data scaling. Parameter scaling is often studied in decoder-only Transformers~\citep{kaplan_scaling_2020, hoffmann_training_2022}, with newer works addressing small and efficient models~\citep{hu_minicpm:_2024, clark_unified_2022}. These studies establish power-law relationships between loss and model size or compute (Equation~\ref{eq:basic_scaling}). In parallel, \textbf{data scaling} research has proposed laws for optimizing mixtures~\citep{ye_data_2024}, repeated training exposures~\citep{muennighoff_scaling_2023}, vocabulary size~\citep{tao_scaling_2024}, and knowledge capacity~\citep{allen-zhu_physics_2024}.

\noindent \textbf{Pre-training scaling laws} extend beyond language to vision and multimodal settings. Vision models exhibit power-law scaling that saturates at large compute~\citep{zhai_scaling_2022}, while multimodal models demonstrate competition-to-synergy transitions as scale increases~\citep{aghajanyan_scaling_2023}.

\noindent \textbf{Post-training scaling} captures fine-tuning and transfer learning behaviors. Transfer scaling shows larger pre-trained models yield better generalization with limited downstream data~\citep{hernandez_scaling_2021}. Recent works propose scaling laws for PEFT~\citep{zhang_when_2024}, downstream loss prediction~\citep{chen_scaling_2024}, and early stopping~\citep{lin_selecting_2024}.

\noindent \textbf{Inference scaling} explores compute-efficient strategies during model deployment. Adaptive test-time compute~\citep{chen_are_2024, brown_large_2024} and retrieval augmentation~\citep{shao_scaling_2024} allow small models to rival larger ones. Inference-specific scaling laws characterize the tradeoff between sampling cost and performance~\citep{wu_inference_2024}.

\noindent \textbf{Efficient model scaling} addresses sparsity, quantization, and distillation. Sparse and MoE models provide multiplicative efficiency gains~\citep{krajewski_scaling_2024}, while pruning and quantization laws enable compute-aware compression~\citep{chen2024scaling, cao_scaling_2024}.

\noindent \textbf{Scaling behavior in reinforcement learning} (RL) diverges from language or vision tasks. In single-agent RL, performance scales sublinearly with model size and environment interaction~\citep{hilton_scaling_2023}. Horizon length, rather than task difficulty, determines scaling efficiency. In multi-agent games, predictable scaling laws govern compute-to-performance relationships, but generalization to complex domains like Chess or Go remains limited~\citep{neumann_scaling_2023}. Meanwhile, graph neural networks (GNNs) lack stable scaling laws; despite self-supervised loss improving with more data, downstream performance often fluctuates unpredictably~\citep{ma_neural_2024}.

Finally, the taxonomy captures two outer branches: \textbf{commendations}, such as practical data laws and compression-aware training~\citep{liu_regmix:_2024}, and \textbf{criticisms}, which question the generalizability and reproducibility of scaling laws~\citep{sorscher_beyond_2023, diaz_scaling_2024}. Detailed discussion on these scaling law studies are provided in Appendix~\ref{appx:details}. 

In the next section, we formulate key research questions (mapping between the taxonomy and research questions highlighted in Table~\ref{tab:taxonomy_map}) derived from these studies and present practical guidelines for leveraging scaling laws in real-world model development.

\input{tables/taxonomy_mapping}

\section{Research questions and guidelines}
\label{appx:rq}

Grounded in the taxonomy of neural scaling laws (Figure~\ref{fig:lit_surv}), we identify key research questions spanning six dimensions: \textit{model scaling}, \textit{architectural bottlenecks}, \textit{inference scaling}, \textit{data scaling}, \textit{post-training strategies}, and \textit{efficient model design}. For each, we synthesize multiple studies to extract overarching patterns, identify conflicting evidence, and propose actionable guidelines for researchers and practitioners navigating large-scale model development.


\subsubsection*{RQ1. Importance on model and pre-training data size on performance. [taxonomy: model scaling $\rightarrow$ pre-training]}

\citet{kaplan_scaling_2020} established a power-law relationship:
\begin{equation}
\small
L(N,D) = \left[\left(\frac{N_c}{N}\right)^{\frac{\alpha_N}{\alpha_D}} + \frac{D_c}{D}\right]^{\alpha_D}, \quad D \propto N^{0.74}.
\end{equation}

\citet{hoffmann_training_2022} refined this into a compute-optimal formulation:
\begin{equation}
\small
\label{eq:chinchilla}
L(N,D) = \frac{A}{N^\alpha} + \frac{B}{D^\beta} + E, \quad D \propto N.
\end{equation}

Recent research has challenged linear extrapolations. \citet{muennighoff_scaling_2023} and \citet{sardana_beyond_2024} showed that training small models longer can outperform larger models, especially under constrained data. \citet{caballero_broken_2023} proposed Broken Neural Scaling Laws (BNSL):
\begin{equation}
\small
L(N, D) = \begin{cases} aN^{-\alpha} + bD^{-\beta}, & N < N_c \\ cN^{-\alpha'} + dD^{-\beta'}, & N \geq N_c \end{cases}
\end{equation}

\begin{tcolorbox}[title=Synthesis and guidelines, colback=gray!5, colframe=gray!50!black]
\begin{itemize}[noitemsep, topsep=0pt, left=0pt]\small
    \item Model scaling success depends not only on size but also on training strategy, data quality, and saturation thresholds.
    \item Practitioners should allocate compute across parameters, data, and training duration based on observed inflection points. Use Kaplan/Chinchilla scaling when data is abundant; otherwise, extend training epochs or adopt data-efficient curricula (see Figure~\ref{fig:training_strategy}).
\end{itemize}
\end{tcolorbox}

\subsubsection*{RQ2. Scaling behaviors for different neural architectures. [taxonomy: model scaling $\rightarrow$ pre-training $\rightarrow$ architecture]}

According to \citet{tay_scaling_2022}, the vanilla Transformer consistently demonstrates superior scaling properties ($P \propto  C^\alpha$, where \( P \) is the performance metric, \( C \) represents compute, and \( \alpha \) are fitting parameters) compared to other architectures, even though alternative designs might perform better at specific sizes. Architectural bottlenecks manifest differently across these designs. For instance, linear attention models like Performer and Lightweight Convolutions show inconsistent scaling behavior, while ALBERT demonstrates negative scaling trends. This finding helps explain why most LLMs maintain relatively standard architectures rather than adopting more exotic variants. Furthermore, ~\citet{zhai_scaling_2022} revealed that ViT reveals that these models exhibit \textit{double saturation}, where performance plateaus at both very low and very high compute levels, suggesting architectural limitations specific to the vision domain (Equation \ref{eq:vit}). However, as shown by \citet{li_are_2024}, simply scaling up vision encoders in multimodal models does not consistently improve performance, indicating that architectural scaling benefits are not uniform across modalities.
\begin{align}
\small
    E = a (C + d)^{-b} + c,
    \label{eq:vit}
\end{align}
where \( E \) denotes downstream error, \( C \) represents compute, and \( a, b, c, d \) are fitting parameters.

\begin{tcolorbox}[title=Synthesis and guidelines, colback=gray!5, colframe=gray!50!black]
\begin{itemize}[noitemsep, topsep=0pt, left=0pt]\small
    \item Architectural bottlenecks vary across domains and compute scales. Transformer inductive biases generalize best under scale.
    \item Use architectures with proven scaling profiles (e.g., vanilla Transformer) unless task-specific benefits outweigh risks. For multimodal or domain-specialized setups, consult scaling behavior across compute ranges (Figure~\ref{fig:training_strategy}).
\end{itemize}
\end{tcolorbox}

\subsubsection*{RQ3. Data strategies for performance scaling. [taxonomy: data scaling]}

~\citet{ye_data_2024} proposed an exponential model for data mixing:
\begin{equation}
\small
L_i(r_{1...M}) = c_i + k_i \exp\left(\sum_{j=1}^M t_{ij}r_j\right),
\end{equation}
while ~\citet{liu_regmix:_2024} and ~\citet{kang_autoscale:_2024} developed proxy models (REGMIX, AUTOSCALE) to pre-optimize mixtures. The Domain-Continual Pretraining (D-CPT) law~\citep{que_d-cpt_2024} provides a theoretical grounding on optimal mixture  ratio between general and domain-specific data :
\begin{equation}
\small
L(N, D, r) = E + \frac{A}{N^\alpha} + \frac{B \cdot r^\eta}{D^\beta} + \frac{C}{(r + \epsilon)^\gamma},
\end{equation}
where $N$ represents the number of model parameters, $D$ is the dataset size, $r$ is the mixture ratio,
$E, A, B, C, \alpha, \beta, \gamma, \eta, \epsilon$ are fitting parameters.

\begin{tcolorbox}[title=Synthesis and guidelines, colback=gray!5, colframe=gray!50!black]
\begin{itemize}[noitemsep, topsep=0pt, left=0pt]\small
    \item Model performance is sensitive to data heterogeneity, mixture ratios, and interaction effects -- especially in multi-domain or continual settings.

    \item Replace manual corpus aggregation with predictive data mixing. Use D-CPT law when adapting to specific domains. Figure~\ref{fig:training_strategy} outlines strategy paths based on data availability and domain constraints.
\end{itemize}
\end{tcolorbox}

\subsubsection*{RQ4. Test-time scaling for better scaling efficiency. [taxonomy: model scaling $\rightarrow$ inference scaling]}

Recent research examining the relationship between test-time computation and model size scaling has revealed key insights. \citet{brown_large_2024} proposed that repeated sampling during inference significantly enhances model performance, with coverage \( C \) (fraction of problems solved) following an exponentiated power law relationship with the number of samples \( k \), $\log(C) = ak^{-b}$, where \(a,b\) are fitting parameters. Further exploration by \citet{wu_inference_2024} suggested that employing sophisticated test-time computation strategies (such as iterative refinement or tree search) with smaller models may be more cost-effective than using larger models with simple inference methods. Their work establishes a relationship between inference computational budget and optimal model size for compute-efficient inference, expressed as $\log_{10}(C) = 1.19\log_{10}(N) + 2.03$.

\begin{tcolorbox}[title=Synthesis and guidelines, colback=gray!5, colframe=gray!50!black]
\begin{itemize}[noitemsep, topsep=0pt, left=0pt]\small
    \item Inference scaling offers a complementary path to performance, particularly where model reuse is desired but compute cost must remain low.
    
    \item Use adaptive compute, retrieval augmentation, or tree search for high-value queries. Integrate test-time scaling laws into deployment workflows (Figure~\ref{fig:inference_strategy}).
\end{itemize}
\end{tcolorbox}

\subsubsection*{RQ5. Scaling behaviors of model fine-tuning. [taxonomy: model scaling $\rightarrow$ post-training scaling]}

Fine-tuning scaling reflects how pre-trained models adapt across tasks and domains. \citet{hernandez_scaling_2021} introduced a transfer scaling law based on effective data transferred $D_t$:
\begin{equation}
\small
D_t(D_f,N) = k(D_f)^\alpha(N)^\beta,
\end{equation}
while \citet{lin_selecting_2024} refined this with a rectified law:
\begin{equation}
\small
L(D) = \frac{B}{D_t + D^\beta} + E,
\end{equation}
modeling diminishing returns from fine-tuning beyond a pre-learned threshold. In vision, \citet{abnar_exploring_2021} linked downstream error to upstream error:
\begin{equation}
\small
e_{DS} = k(e_{US})^a + c,
\end{equation}
and \citet{mikami_scaling_2021} connected downstream accuracy to synthetic pretraining data size:
\begin{equation}
\small
e_{DS} = aD^{-\alpha} + c.
\end{equation}

\textit{FLOPS to Loss to Performance} (FLP) method~\citep{chen_scaling_2024} predicted downstream performance from pretraining FLOPs, and \citet{zhang_when_2024} showed LoRA scales nonlinearly under PEFT:
\begin{equation}
\small
\hat{L}(X, D_f) = A \times \frac{1}{X^\alpha} \times \frac{1}{D_f^\beta} + E.
\end{equation}

\begin{tcolorbox}[title=Synthesis and guidelines, colback=gray!5, colframe=gray!50!black]
\begin{itemize}[noitemsep, topsep=0pt, left=0pt]\small
    \item Transferability scales with both model size and pretraining loss, but task difficulty, data availability, and adaptation type mediate returns.

    \item Use FLP or rectified laws to estimate post-training gains. Prefer PEFT for low-resource settings; switch to full fine-tuning when compute permits. For domain adaptation, apply D-CPT strategies (Figure~\ref{fig:training_strategy}).
\end{itemize}
\end{tcolorbox}

\subsubsection*{RQ6. Scaling efficiency and performance for sparse and efficient models. [taxonomy: model scaling $\rightarrow$ model compression]}

As the demand for resource-efficient models grows, sparse architectures such as pruned networks and MoEs have emerged as promising alternatives to dense Transformers. These models aim to preserve the performance benefits of scale while reducing compute and memory overhead. \citet{frantar_scaling_2023} proposed a general sparse scaling law showing that sparsity acts as a multiplicative efficiency factor rather than changing the fundamental scaling behavior:
\begin{equation}
\small
L(S,N,D) = (a_S(1-S)^{b_S} + c_S) \cdot \left(\frac{1}{N}\right)^{b_N} + \left(\frac{a_D}{D}\right)^{b_D} + c,
\label{eq:sparse_law}
\end{equation}
where \(S\) is sparsity, \(N\) is the number of non-zero parameters, and \(D\) is dataset size. In MoE models, where only a subset of parameters is activated per input, \citet{clark_unified_2022} proposed a loss scaling relationship incorporating both model size and expert count:
\begin{equation}
\small
\log L = a \log N + b \log E + c \log N \cdot \log E + d,
\label{eq:clark_moe}
\end{equation}
with \(E\) denoting the expansion factor. This formulation was extended by \citet{yun_toward_2024} to include dataset size:

{\small
\begin{align}
    \log L(N,D,E) &= \log\left( \frac{a}{N^\alpha} + \frac{b}{E^\beta} + \frac{c}{D^\gamma} + f \right) \notag\\ &+ d \log N \log E
\end{align}
}%
These results emphasize that scaling MoEs effectively requires balancing expert granularity with sufficient training data. Toward this, \citet{krajewski_scaling_2024} introduced a granularity parameter \(G\) to refine the Chinchilla-style formulation:
\begin{equation}
\small
\mathcal{L}(N,D,G) = c + \left(\frac{g}{G^\gamma} + a\right)\frac{1}{N^\alpha} + \frac{b}{D^\beta}.
\label{eq:granularity}
\end{equation}

In parallel, structured pruning approaches have been formalized through the \(P^2\) law~\citep{chen2024scaling}, which relates post-pruning loss to pre-pruning model size \(N_0\), pruning ratio \(\rho\), and post-training token count \(D\):
\begin{equation}
\small
L(N_0, D, \rho, L_0) = L_0 + \left( \frac{1}{\rho} \right)^\gamma \left( \frac{1}{N_0} \right)^\delta \left( \frac{N_C}{N_0^\alpha} + \frac{D_C}{D^\beta} + E \right),
\label{eq:p2_law}    
\end{equation}
where \( L_0 \) is the uncompressed model loss, \( \rho \) is the pruning rate, \( N_0 \) is the pre-pruning model size, \( D \) represents the number of post-training tokens, and \( N_C, D_C, E, \alpha, \beta, \gamma \) are fitting parameters.

\begin{tcolorbox}[title=Synthesis and guidelines, colback=gray!5, colframe=gray!50!black]
\begin{itemize}[noitemsep, topsep=0pt, left=0pt]\small
    \item Sparse models are scaling-compliant but require careful routing (MoE) and token-budget tuning (pruning) to outperform dense counterparts.
    
    \item Use MoEs for general-purpose LLMs under compute limits. Apply pruning for deployment constraints. For efficient inference, refer to Figure~\ref{fig:inference_strategy}.
\end{itemize}
\end{tcolorbox}

\subsubsection*{RQ7. Model scaling with low-precision quantization. [taxonomy: model scaling $\rightarrow$ model compression $\rightarrow$ quantization ]}

According to~\citet{dettmers_case_2023}, 4-bit precision appears to be the optimal sweet spot for maximizing model performance while minimizing model size. Additionally, research on scaling with mixed quantization~\citep{cao_scaling_2024}, demonstrated that larger models can handle higher quantization ratios while maintaining performance, following an exponential relationship where larger models require exponentially fewer high-precision components to maintain a given performance level. \citet{kumar_scaling_2024} developed a unified scaling law (Equation \ref{eq:unified_scaling_quant}) that predicts both training and post-training quantization effects. It further suggests that effects of quantizing weights, activations, and attention during training are independent and multiplicative. 
\begin{equation}
\small
    L(N, D, P_w, P_a, P_{\text{kv}}, P_{\text{post}}) =  AN_{\text{eff}}^{-\alpha} + BD^{-\beta} + E + \delta_{PTQ},
\label{eq:unified_scaling_quant}
\end{equation}
where $P_w,P_a,P_{kv}$ denote training precision of weights, activations and attentions, respectively, $P_{post}$ denote end-time weight-precision, $\delta_{PTQ}$ denotes loss due to post training quantization, and \(\alpha,\beta\) are fitting parameters.

\begin{figure*}
\centering
    \subfloat[Training strategies]{\includegraphics[width=0.6\linewidth]{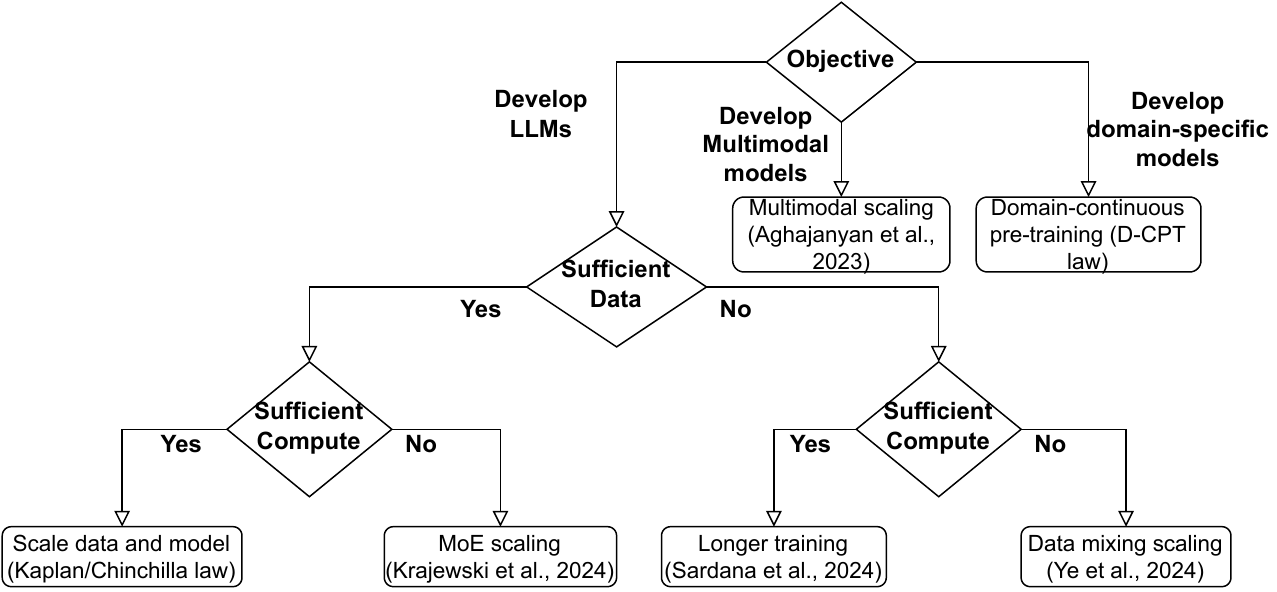}\label{fig:training_strategy}}
    \subfloat[Inference strategies]{\includegraphics[width=0.4\linewidth]{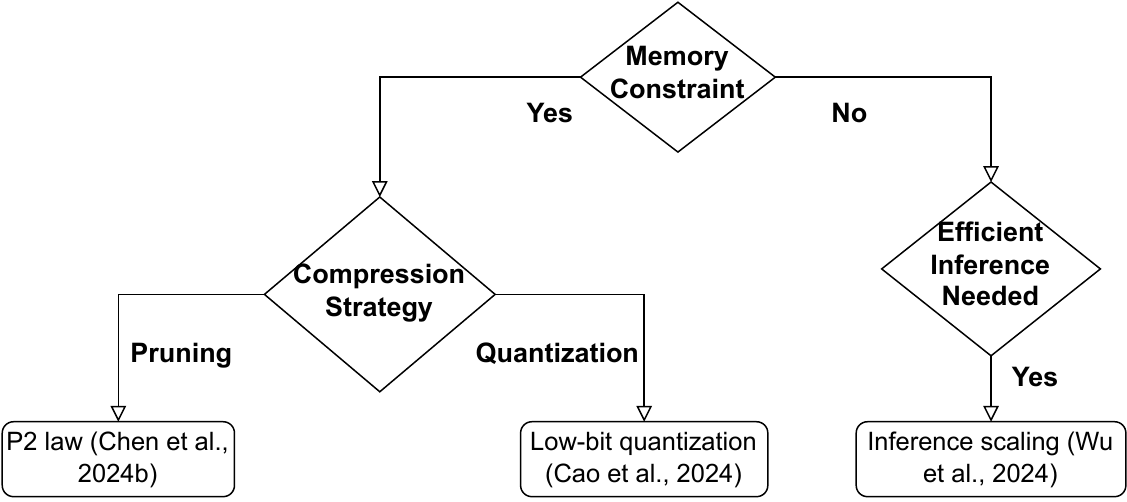} \label{fig:inference_strategy}}
    \caption{Practical roadmap summarizing training and inference strategies grounded in our eight research questions and taxonomy branches. (a) Training scaling strategies can be utilized for pre-training or fine-tuning unimodal and multimodal foundational and domain-adapted models. (b) Post-training inference strategies can be followed to ensure that the model is utilized efficiently for the downstream applications.}
    \label{fig:strategy}
\end{figure*}

\begin{tcolorbox}[title=Synthesis and guidelines, colback=gray!5, colframe=gray!50!black]
\begin{itemize}[noitemsep, topsep=0pt, left=0pt]\small
    \item Scaling-aware quantization reduces memory while preserving performance. Larger models generalize better to low precision.
    
    \item Apply mixed-precision for inference. Use quantization-aware training for smaller models. Refer to post-training strategies (Figure~\ref{fig:inference_strategy}) to guide compression.
\end{itemize}
\end{tcolorbox}

\subsubsection*{RQ8. Beyond modalities: scaling for multimodal models. [taxonomy: model scaling $\rightarrow$ multimodal models]}
Multimodal scaling behavior builds upon, but does not replicate, unimodal trends. \citet{henighan_scaling_2020} first proposed multimodal scaling using $L(x) = Ax^{-\alpha} + B$, where $x$ represents model size, data, or compute. \citet{alabdulmohsin_revisiting_2022} refined this into a more flexible sigmoid-like form:
\begin{equation}
\small
\frac{L_x - L_\infty}{(L_0 - L_x)^\alpha} = \beta x^c,
\end{equation}
allowing transitions across saturation regimes. \citet{aghajanyan_scaling_2023} observed that smaller multimodal models exhibit competition between modalities, while larger models cross a ``competition barrier'' and become synergistic. They proposed a bimodal generalization of the Chinchilla law:
\begin{equation}
\small
\begin{split}
\mathcal{L}(N, D_i, D_j) = \left[\frac{\mathcal{L}(N, D_i) + \mathcal{L}(N, D_j)}{2}\right] \\
 - C_{i,j} + \frac{A_{i,j}}{N^{\alpha_{i,j}}} + \frac{B_{i,j}}{|D_i| + |D_j|^{\beta_{i,j}}},
\end{split}
\end{equation}
where $C_{i,j}$ captures the degree of positive interaction between modalities $i$ and $j$.

\begin{tcolorbox}[title=Synthesis and guidelines, colback=gray!5, colframe=gray!50!black]
\begin{itemize}[noitemsep, topsep=0pt, left=0pt]\small
    \item Multimodal scaling is governed by modality alignment and architectural balance more than raw model size. 
    
    \item Ensure models are sufficiently large to benefit from synergy across modalities. Prioritize modality balance in architecture and high-quality aligned datasets over isolated scaling. Refer to Figure~\ref{fig:training_strategy} when designing multimodal pretraining pipelines.
\end{itemize}
\end{tcolorbox}

\begin{tcolorbox}[title=Cross-RQ synthesis, colback=blue!5, colframe=blue!50!black]
\begin{itemize}[noitemsep, topsep=0pt, left=0pt]\small
    \item Data-efficient scaling (RQ1, RQ3, RQ5) consistently beats brute-force model expansion, as shown in ~\citet{hu_minicpm:_2024,sardana_beyond_2024}.
    \item Architectural innovations (RQ2, RQ6) tend to scale poorly unless paired with precise training heuristics (e.g., expert routing in MoEs).
    \item Inference-aware scaling (RQ4, RQ7) enables small models to rival larger ones but is rarely included in current scaling laws - a key research gap.
\end{itemize}
\end{tcolorbox}
While the research questions synthesized above highlight the strengths and practical applications of neural scaling laws, they also expose several limitations, especially in their generalizability, reliability under constraints and applicability to modern model designs. In the next section, we critically examine these limitations and discuss the foundational assumptions that may no longer hold as models evolve.

\section{Criticisms of scaling laws}
\label{sec:criticism}

\citet{diaz_scaling_2024} challenged the generalizability of neural scaling laws, arguing that they fail in diverse real-world AI applications. They argued that scaling laws do not always hold when AI models serve heterogeneous populations with conflicting criteria for model performance. Larger datasets inherently reflect diverse communities, making it difficult to optimize a single model for all users. Similar to issues in multilingual AI, increasing data diversity often leads to performance degradation rather than improvement. Universal evaluation metrics are inadequate for capturing these complexities, potentially reinforcing biases against underrepresented groups. The authors further argued that smaller, localized AI models may be more effective for specific communities, highlighting the need to move beyond one-size-fits-all scaling assumptions.

Beyond dataset expansion, data pruning contradicts traditional scaling laws by demonstrating that performance improvements do not always require exponentially more data. Strategic pruning achieves comparable or superior results with significantly fewer training samples~\citep{sorscher_beyond_2023}. Not all data contributes equally, and selecting the most informative examples enables more efficient learning. Experimental validation on CIFAR-10, SVHN, and ImageNet shows that careful dataset curation can surpass traditional power-law improvements, questioning the necessity of brute-force scaling.

Despite their significant impact, many studies on scaling laws suffer from limited reproducibility (see Table~\ref{tab:reproducibility} in Appendix~\ref{appx:reproducibility}) due to proprietary datasets, undisclosed hyperparameters, and undocumented training methodologies. The inability to replicate results across different computing environments raises concerns about their robustness. Large-scale experiments conducted by industry labs often depend on private infrastructure, making independent verification challenging. This lack of transparency undermines the reliability of scaling law claims and highlights the urgent need for open benchmarks and standardized evaluation frameworks to ensure reproducibility. Furthermore, the field's avoidance of rigorous scaling exponent analysis constitutes a critical oversight. While exponents indeed vary across models, datasets, and hyperparameters, this variability demands investigation rather than dismissal. This deliberate analytical gap undermines confidence in extrapolation claims and raises questions about whether observed scaling behaviors represent genuine properties or experimental artifacts.

\section{Beyond Scale: Future Directions for Practical and Sustainable AI}
While neural scaling laws have provided valuable insights into model performance, their current formulations often fail to account for recent advancements in architecture, data efficiency, and inference strategies. 
The following directions highlight key areas where scaling laws should be adapted to improve their predictive power and practical utility.




\paragraph{Reframing scaling laws for real-world constraints.} Future scaling laws must account for compute budgets, hardware latency, and energy consumption. This includes integrating training–inference trade-offs, evaluating real-world performance under quantization or pruning, and predicting effectiveness across resource-constrained environments.
\paragraph{Designing for \textit{downscaling}.} Rather than building ever-larger models, the field should invest in scaling laws for \textit{small} language models trained with optimal data, sparsity, and inference strategies. The emergence of 1-3B parameter models that rival 13B+ models~\citep{hu_minicpm:_2024} highlights the viability of compact yet performant systems.
\paragraph{Multi-objective scaling optimization.} Current scaling laws often predict accuracy at scale but ignore trade-offs between accuracy, compute, and robustness. Future work should develop \textit{multi-objective scaling frameworks} that balance these factors to guide architecture and dataset design more holistically.
\paragraph{Inference-aware and modular scaling laws.} Traditional scaling laws assume fixed inference procedures. However, our synthesis in \textbf{RQ4} and \textbf{RQ7} shows that test-time compute allocation via sampling, retrieval, or routing can drastically affect performance. Future scaling formulations should modularize inference and allow flexible compute allocation per task or query.
\paragraph{Data quality over quantity.} Instead of expanding datasets indiscriminately, laws like REGMIX~\citep{liu_regmix:_2024} and D-CPT~\citep{que_d-cpt_2024} emphasize optimized data composition. Future models should prioritize informative examples and track dataset efficiency across tasks.
\paragraph{Towards accessible and sustainable AI.} Large models are inaccessible to many research groups. Downscaling informed by scaling laws ensures that smaller labs and edge deployments can still benefit from state-of-the-art performance. Ultimately, the future of neural scaling is not just bigger models, but \textit{better modeling choices at every scale}.

\section{Conclusion}
This survey provided a comprehensive analysis of neural scaling laws, exploring their theoretical foundations, empirical findings, and practical implications. It synthesized insights across various modalities, including language, vision, multimodal, and reinforcement learning, to uncover common trends and deviations from traditional power-law scaling. While early research established predictable relationships between model size, dataset volume, and computational resources, more recent studies have shown that these relationships are not universally applicable. Sparse architectures, retrieval-augmented models, and domain-specific adaptations often exhibit distinct scaling behaviors, challenging the notion of uniform scalability. Furthermore, advancements in fine-tuning, data pruning, and efficient inference strategies have introduced new perspectives on compute-optimal scaling. Despite their significance, scaling laws remain an evolving area of research, requiring further refinement to address real-world deployment challenges and architectural innovations.

\newpage
\section*{Limitations}
While this survey provides a broad synthesis of neural scaling laws, it primarily focuses on model size, data scaling, and compute efficiency. Other important aspects, such as hardware constraints, energy consumption, and the environmental impact of large-scale AI training, are not deeply explored. Another limitation is the reliance on prior empirical findings, which may introduce variability due to differing experimental setups and proprietary datasets. Without access to fully reproducible scaling law experiments, some conclusions remain dependent on the methodologies employed in original studies.

\section*{Ethical Considerations}
Scaling laws, while effective in optimizing AI performance, can also raise issues of accessibility and fairness. The development of increasingly large models favors institutions with substantial computational resources, creating a divide between well-funded research groups and smaller organizations. Furthermore, as scaling laws often assume uniform data utility, they may amplify biases present in large-scale datasets, potentially leading to skewed outcomes in real-world applications. Ethical concerns also arise from the energy-intensive nature of training large models, contributing to environmental concerns. Addressing these issues requires more inclusive AI development strategies, ensuring that scaling laws consider broader societal impacts rather than focusing solely on performance optimization.

\bibliography{custom}

\clearpage

\appendix

\tableofcontents
\addtocontents{toc}{\protect\setcounter{tocdepth}{2}} 

\section{Fitting and validating scaling laws}
\label{appx:validation}

Fitting scaling laws involves several key methodological choices that can significantly impact the final results and conclusions. The choice of optimization approach, loss function, initialization strategy, and validation method all play crucial roles in determining the reliability and reproducibility of scaling law studies.

\subsection{Optimization methods}
The most common approaches for fitting scaling laws involve non-linear optimization algorithms like BFGS (Broyden-Fletcher-Goldfarb-Shanno) (used by ~\citet{frantar_scaling_2023}), L-BFGS (used by ~\citet{tao_scaling_2024}) and least squares (used by ~\citet{caballero_broken_2023}). Some studies~\citep{covert2024scalinglawsvalueindividual, hashimoto21a} also use optimizers like Adam or Adagrad, though these may be less suitable for scaling law optimization due to their data-hungry nature and assumptions about gradient distributions.

\subsection{Loss functions and objectives}
Several loss functions are commonly used for fitting scaling laws:
\begin{itemize}[leftmargin=*,noitemsep,topsep=0pt]
    \item \textbf{Mean squared error (MSE)}: Emphasizes larger errors due to quadratic scaling (used by ~\citet{ghorbani2021scaling}).
\item \textbf{Mean absolute error (MAE)}: Provides more robust fitting less sensitive to outliers (used by ~\citet{hilton_scaling_2023}).
\item \textbf{Huber loss}: Combines MSE's sensitivity to small errors with MAE's robustness to outliers (used by ~\citet{hoffmann_training_2022}).
\end{itemize}

\subsection{Initialization strategies}
The initialization of scaling law parameters proves to be critically important for achieving good fits. Common approaches include grid search over parameter spaces~\citep{aghajanyan_scaling_2023}, random sampling from parameter ranges~\citep{frantar_scaling_2023}, and multiple random restarts to avoid local optima~\citep{caballero_broken_2023}.


\subsection{Validation methods}
It is hugely important to understand if the scaling law fit achieved is accurate and valid. Most of the papers surveyed lack in validating their fits. Several approaches can help validating the effectiveness of scaling law fits. Statistical methods like computing confidence intervals can act as a goodness-of-fit metric~\citep{alabdulmohsin_revisiting_2022}. Furthermore, researchers can perform out-of-sample testing by extrapolation to larger scales~\citep{hoffmann_training_2022}.

\subsection{Limitations of fitting techniques}
\citet{li_misfitting_2024} revealed several critical methodological considerations in fitting scaling laws. Different optimizers can converge to notably different solutions even with similar initializations, underscoring the need for careful justification of optimizer choice. Similarly, the analysis showed that different loss functions can produce substantially different fits when working with real-world data containing noise or outliers, suggesting that loss function selection should be guided by specific data characteristics and desired fit properties. Perhaps most importantly, the paper demonstrated that initialization can dramatically impact the final fit, with some methods exhibiting high sensitivity to initial conditions. Together, these findings emphasize the importance of thorough methodology documentation across all aspects of the fitting process - from optimizer selection and loss function choice to initialization strategy - to ensure reproducibility and reliability in scaling law studies.

\section{Detailed scaling laws}
\label{appx:details}

\subsection{Scaling laws of language models}
\citet{kaplan_scaling_2020} suggested that larger LMs improve performance by reducing loss through power-law scaling. However, this view evolved when studies showed that many large models were undertrained, and data scaling plays an equally crucial role in compute efficiency~\citep{hoffmann_training_2022}. More recent breakthroughs challenged traditional scaling assumptions. Broken Neural Scaling Law (BNSL) introduced non-monotonic trends, meaning that model performance can sometimes worsen before improving, depending on dataset thresholds and architectural bottlenecks~\citep{caballero_broken_2023}. Another exciting development came from small LMs, where optimized training strategies, such as a higher data-to-parameter ratio and adaptive learning schedules, enable models ranging from 1.2B to 2.4B parameters to rival significantly larger 7B-13B models~\citep{hu_minicpm:_2024}. These findings reshape the fundamental assumptions of scaling laws, proving that strategic training can outperform brute-force model expansion. 

\input{tables/scaling_laws1}

\subsection{Scaling laws in other modalities}
\input{tables/scaling_laws2}

In computer vision, ViTs exhibit power-law scaling when model size, compute, and data grow together, but their performance plateaus at extreme compute levels, with noticeable gains only when trained on datasets exceeding 1B images~\citep{zhai_scaling_2022}. Meanwhile, studies on scaling law extrapolation revealed that while larger models generally scale better, their efficiency declines at extreme sizes, requiring new training strategies to maintain performance~\citep{alabdulmohsin_revisiting_2022}. In multimodal learning, an interesting phenomenon called the ``competition barrier'' has been observed where at smaller scales different input modalities compete for model capacity, but as models grow, they shift into a synergistic state, enabling accurate performance predictions based on model size and token count~\citep{aghajanyan_scaling_2023}.

However, not all scaling trends align with expectations. Contrary to the assumption that larger is always better, scaling vision encoders in vision-language models can sometimes degrade performance, highlighting the fact that data quality and modality alignment are more critical than brute-force scaling~\citep{li_are_2024}. These findings collectively emphasize that scaling laws are domain-dependent -- optimal scaling strategies require a careful balance between compute efficiency, dataset quality, and architecture rather than simply increasing model size. Table~\ref{tab:scaling_laws1} summarizes the scaling laws of pre-trained models for language and other modalities.

\subsection{Scaling laws for domain adaptation}
Pre-training and fine-tuning techniques have accelerated the adoption of large-scale neural models, yet the extent to which these models transfer across tasks and domains remains a key research question tied to scaling principles. Studies show that transfer learning follows a power-law where pre-training amplifies fine-tuning effectiveness, especially in small data regimes. Even with limited downstream data, larger models benefit significantly from pre-training, improving generalization~\citep{hernandez_scaling_2021}. In vision, pre-training saturation occurs due to upstream-downstream interactions, rather than just task complexity. Lower network layers quickly specialize in simple tasks, while higher layers adapt to complex downstream objectives~\citep{abnar_exploring_2021}. Similarly, in synthetic-to-real transfer, larger models consistently reduce transfer gaps, enhancing generalization across domains~\citep{mikami_scaling_2021}.

Fine-tuning strategies scale differently depending on dataset size. Parameter-efficient fine-tuning (PEFT) techniques like low-rank adaptation (LoRA)~\citep{hu2021lora} and Prompt-tuning, both are well-suited for small datasets, but LoRA performs best for mid-sized datasets, and full fine-tuning is most effective for large datasets. However, PEFT methods provide better generalization in large models, making them attractive alternatives to full-scale fine-tuning~\citep{zhang_when_2024}.

Scaling laws are also being utilized to accurately predict the fine-tuning performance of models. The FLP method~\citep{chen_scaling_2024} estimates pre-training loss from FLOPs, enabling accurate forecasts of downstream performance, particularly in models up to 13B parameters. Further refinements like FLP-M improve mixed-dataset predictions and better capture emergent abilities in large models. Finally, the Rectified scaling law~\citep{lin_selecting_2024} introduces a two-phase fine-tuning transition, where early-stage adaptation is slow before shifting into a power-law improvement phase. This discovery enables compute-efficient model selection using the ``Accept then Stop'' (AtS) algorithm to terminate training at optimal points.

\input{tables/scaling_laws3}

We summarize these findings in Table~\ref{tab:scaling_laws3}, suggesting that transfer learning is highly scalable, but effective scaling requires precise tuning strategies rather than just increasing model size.

\subsection{Scaling laws for model inference}

Simply scaling up models is not always the best way to improve model performance. ~\citet{chen_are_2024} suggested that more efficient test-time compute strategies can dramatically reduce inference costs while maintaining or even exceeding performance. Instead of blindly increasing LLM calls, they further suggested for allocating resources based on query complexity, ensuring that harder queries receive more compute while simpler ones use fewer resources. The importance of test-time compute strategies becomes even clearer when dealing with complex reasoning tasks. While sequential modifications work well for simple queries, parallel sampling and tree search dramatically improve results on harder tasks. Adaptive compute-optimal techniques have been shown to reduce computational costs by 4$\times$ without degrading performance, allowing smaller models with optimized inference strategies to surpass much larger models~\citep{snell_scaling_2024, brown_large_2024}. Advanced inference approaches, such as REBASE tree search~\citep{wu_inference_2024}, further push the boundaries of efficiency, enabling small models to perform on par with significantly larger ones. \\

Another breakthrough came from retrieval augmented models, where increasing the datastore size consistently improves performance without hitting saturation \citep{shao_scaling_2024}. This allows smaller models to outperform much larger ones on knowledge-intensive tasks, reinforcing that external datastores provide a more efficient alternative to memorizing information in model parameters.

\input{tables/scaling_laws4}

\subsection{Scaling laws for efficient models}
\input{tables/scaling_laws5}

Scaling laws have expanded beyond simple parameter growth, introducing new methods to optimize routing, sparsity, pruning, and quantization for efficient LLM scaling. Routing-based models benefit from optimized expert selection, but their returns diminish at extreme scales, requiring careful expert configuration~\citep{clark_unified_2022}. In contrast, fine-grained MoE models consistently outperform dense transformers, achieving up to 40$\times$ compute efficiency gains when expert granularity is properly tuned~\citep{krajewski_scaling_2024}. However, balancing the number of experts ($E$) is crucial, where models with 4-8 experts offer superior inference efficiency, but require $2.5-3.5\times$ more training resources, making 16-32 expert models more practical when combined with extensive training data~\citep{yun_toward_2024}. Sparse model scaling offers another efficiency boost. Research has demonstrated that higher sparsity enables effective model scaling, allowing $2.15\times$ more parameters at 75\% sparsity, improving training efficiency while maintaining performance~\citep{frantar_scaling_2023}. Additionally, pruning laws ($\text{P}^{2}$ scaling laws) predict that excessive post-training data does not always improve performance, helping optimize resource allocation in pruned models~\citep{chen2024scaling}. \citet{dettmers_case_2023} showed that 4-bit quantization provides the best trade-off between accuracy and model size, optimizing zero-shot performance while reducing storage costs. Larger models tolerate lower precision better, following an exponential scaling law where fewer high-precision components are needed to retain performance~\citep{cao_scaling_2024}. Meanwhile, training precision scales logarithmically with compute budgets, with 7-8 bits being optimal for balancing size, accuracy, and efficiency~\citep{kumar_scaling_2024}. Recent reserach has expanded into distillation as well, developing a mathematical framework that predicts how well a student model will perform based on the student model's size, the teacher model's performance and the compute budget allocation between the teacher and the student~\citep{busbridge2025distillation}.
We summarize these practical insights in Table~\ref{tab:scaling_laws4} for better readability. 

\subsection{Data scaling laws}
\input{tables/scaling_laws6}
Scaling models involves more than just increasing parameters; optimizing data mixtures, training duration, and vocabulary size also plays a crucial role in enhancing performance and efficiency. Data mixing laws allow AI practitioners to accurately predict optimal data compositions before training, leading to 27\% fewer training steps without compromising accuracy~\citep{ye_data_2024}. Techniques like REGMIX optimize data selection using proxy models and regression, reducing compute costs by 90\% compared to manual data selection~\citep{liu_regmix:_2024}. Meanwhile, AUTOSCALE revealed that data efficiency depends on model scale, where high-quality data like Wikipedia helps small models but loses effectiveness for larger models, which benefit from diverse datasets like CommonCrawl~\citep{kang_autoscale:_2024}. For continual learning, the D-CPT Law provided a theoretical framework for balancing general and domain-specific data, guiding efficient domain adaptation and long-term model updates~\citep{que_d-cpt_2024}. Additionally, Chinchilla scaling assumptions were challenged by evidence showing that training models for more epochs on limited data can outperform simply increasing model size~\citep{muennighoff_scaling_2023}. Repeated data exposure remains stable up to 4 epochs, but returns diminish to zero after around 16 epochs, making longer training a more effective allocation of compute resources. Furthermore, the vocabulary scaling law suggested that as language models grow larger, their optimal vocabulary size should increase according to a power law relationship~\citep{tao_scaling_2024}. Finally, knowledge capacity scaling laws established that language models store 2 bits of knowledge per parameter, meaning a 7B model can encode 14B bits of knowledge -- surpassing English Wikipedia and textbooks combined~\citep{allen-zhu_physics_2024}.
Table~\ref{tab:scaling_laws6} summarizes the data scaling laws for developing neural models when data is not available in abundance.

\subsection{Scaling laws for reinforcement learning}

Scaling laws in reinforcement learning (RL) and reward model optimization reveal both similarities and differences with generative modeling. Single-agent RL follows power-law scaling with model size and environment interactions, with optimal scaling exponents between 0.4-0.8 across tasks lower than the 0.5 exponent observed in language models~\citep{hilton_scaling_2023}. RL tasks require orders of magnitude smaller models than generative tasks, correlating with task horizon length, which dictates environment interaction scaling. Task difficulty increases compute needs but does not affect scaling exponents, highlighting horizon length as a key factor in RL scaling efficiency.

In board games like Hex which involves multi-agent RL, \citet{jones_scaling_2021} showed that AlphaZero performance follows predictable scaling trends, with compute requirements increasing 7$\times$ per board size increment for perfect play and 4$\times$ for surpassing random play~\citep{jones_scaling_2021}. 
\citet{neumann_scaling_2023} extended this study to Pentago and ConnectFour, proposing scaling laws which show that player strength scales with network size as $\alpha_N \approx 0.88$, performance with compute as $\alpha_C \approx 0.55$, and optimal network size with compute budget as $\alpha_{\text{opt}} \approx 0.63$~\citep{neumann_scaling_2023}. Larger multi-agent models exhibit higher sample efficiency, though these trends may not generalize to highly complex games like Chess and Go.

Reward model overoptimization in RLHF follows distinct functional forms: Best-of-$n$ (BoN) reward optimization is governed by $d(\alpha_{\text{bon}} - \beta_{\text{bon}} d)$, whereas RL reward optimization follows $d(\alpha_{\text{RL}} - \beta_{\text{RL}} \log d)$, where $d$ represents KL divergence from the initial policy~\citep{gao_scaling_2022}. RL requires higher KL divergence than BoN for optimization, and reward model overoptimization scales logarithmically with model size, while policy size has minimal impact. 
These findings reinforce the importance of balancing compute allocation, environment complexity, and optimization techniques to achieve scalable and efficient RL models.

\input{tables/database_desc1}
\input{tables/database_desc2}

\input{tables/github_link}

\subsection{Scaling laws for sparse autoencoders}

Recent research has established scaling laws for dictionary learning, providing insights into how latent representations and sparsity impact reconstruction error and computational efficiency. Sparse autoencoders with top-$K$ selection follow power-law scaling for reconstruction error (MSE) in terms of the number of latents $n$ and sparsity $k$, though this relationship only holds for small $k$ relative to model dimension~\citep{gao_scaling_2024}. Larger language models require more latents to maintain the same MSE at a fixed sparsity, reinforcing that latent dimensionality must scale with model size for effective reconstruction. Additionally, MSE follows a power-law relationship with the compute used during training, suggesting that efficient scaling strategies must balance sparsity, latent size, and training compute to minimize error effectively. This is reinforced by \citet{circuits_nodate}, showing that feature representations follow predictable scaling trends, where larger models develop richer, more interpretable dictionaries as the number of learned features increases.


\subsection{Scaling laws for graph neural networks}
Unlike in computer vision and natural language processing, where larger datasets typically improve generalization, graph self-supervised learning methods fail to exhibit expected scaling behavior and performance fluctuates unpredictably across different data scales~\citep{ma_neural_2024}. However, self-supervised learning pretraining loss does scale with more training data, but this improvement does not translate to better downstream performance. The scaling behavior is method-specific, with some approaches like InfoGraph showing more stable scaling than others like GraphCL.

\section{Reproducibility of scaling laws papers}
\label{appx:reproducibility}

The reproducibility status of neural scaling law papers presents a mixed landscape in terms of research transparency. We consolidate and provide the links to github code repositories in the Table \ref{tab:reproducibility}. Among the 45 surveyed papers proposing scaling laws, 
22 papers (48.9\%) provided repository links, indicating some level of commitment to open science practices.  However, more than half of the papers still lack basic reproducibility elements, with 29 papers (64.4\%) not sharing training code and 27 papers (60\%) withholding analysis code. This comprehensive survey suggests that while there is a growing trend toward reproducibility in neural scaling law research, there remains substantial room for improvement in establishing standard practices for code sharing and result verification.

\end{document}

%% file: tables/survey_differences.tex
\begin{table}[!t]
  \centering
  \resizebox{\columnwidth}{!}{
  \begin{tabular}{p{6cm}|c|c|c}
  \cline{1-4}
  \multicolumn{1}{c|}{\textbf{Category}} & \textbf{~\citet{choshen2024hitchhikersguidescalinglaw}} & \textbf{~\citet{li_misfitting_2024}} & \textbf{Ours} \\
  \cline{1-4}
    Covers neural scaling laws broadly & Yes & No & Yes \\
    Discusses fitting methodologies & Yes & Yes & Yes \\
    Analyzes architectural considerations & No & Limited & Yes \\
    Includes data scaling and pruning & No & Limited & Yes \\
    Explores inference scaling & No & Limited & Yes \\
    Considers domain-specific scaling & No & No & Yes \\
    Provides practical guidelines & Yes & Yes & Yes \\
    Critiques limitations of scaling laws & Limited & Yes & Yes \\
    Proposes future research directions& Limited & Yes & Yes \\
    \cline{1-4}
    \end{tabular}
    }

  \caption{Key differences between our survey and existing surveys on neural scaling laws~\citep{choshen2024hitchhikersguidescalinglaw, li_misfitting_2024}.}
  \label{tab:differences}%
  \vspace{-5mm}
\end{table}%

%% file: tables/taxonomy_mapping.tex
\begin{table}[t]
\centering
\small
\begin{tabular}{p{3.6cm} | p{2.8cm}}
\cline{1-2}
\textbf{Taxonomy Node} & \textbf{Addressed RQs} \\
\cline{1-2}
Model scaling & RQ1, RQ2, RQ8 \\
Data scaling & RQ3 \\
Post-training scaling & RQ5 \\
Inference scaling & RQ4 \\
Efficient and compressed model scaling & RQ6, RQ7 \\
\cline{1-2}
\end{tabular}
\caption{Mapping taxonomy categories to relevant research questions.}
\label{tab:taxonomy_map}
\end{table}

%% file: tables/scaling_laws1.tex
\begin{table*}[!htb]
    \centering
    \scalebox{0.9}{
    \begin{tabular}{c p{3.5cm} p{5.6cm} p{5cm} }
        \cline{1-4}
        \textbf{Modality} & \textbf{Paper} & \textbf{Key insights} & \textbf{Applicability} \\
        \cline{1-4}
        \multirow{14}{*}{Language} & \multirow{4}{*}{\citet{kaplan_scaling_2020}} & Larger models are more sample-efficient, needing fewer training examples to generalize well. & Predicts model loss decreases with increasing parameters, used in early LMs like GPT-3.\\
        \cdashline{2-4}
        & \multirow{3}{*}{\citet{hoffmann_training_2022}} & The best performance comes from balancing model size and data, rather than just increasing parameters. & Balances compute, model size, and dataset size for optimal efficiency, as seen in Chinchilla.\\
        \cdashline{2-4}
        & \multirow{3}{*}{\citet{caballero_broken_2023}} & Performance does not always improve smoothly; there are inflection points where scaling stops working. & Identifies phase transitions, minimum data thresholds, and unpredictability in scaling behavior. \\
        \cdashline{2-4}
        & \multirow{4}{*}{\citet{hu_minicpm:_2024}} & Smaller models with better training can rival much larger models. & Demonstrates that smaller models with optimized training can outperform larger undertrained models.\\
        \cline{1-4}
        \multirow{4}{*}{Vision} & \multirow{4}{*}{\citet{zhai_scaling_2022}} & ViTs follow power-law scaling but plateau at extreme compute levels, with benefits primarily seen in datasets >1B images. & Image classification, object detection, large-scale vision datasets. \\
        \cline{1-4}
        \multirow{10}{*}{Multimodal} & \multirow{4}{*}{\citet{aghajanyan_scaling_2023}} & Multimodal models experience competition at smaller scales but transition into synergy as model and token count grow. & Multimodal learning, mixed-modal generative models, cross-domain AI. \\
        \cdashline{2-4}
        & \multirow{5}{*}{\citet{li_are_2024}} & Scaling vision encoders in vision-language models does not always improve performance, reinforcing the importance of data quality over raw scaling. & Vision-language models, image-text alignment, multimodal scaling challenges. \\
        \cline{1-4}
    \end{tabular}
    }
    \caption{Critical neural scaling laws for language, vision and multimodal models.}
    \label{tab:scaling_laws1}
\end{table*}

%% file: tables/scaling_laws2.tex
\begin{table*}[!tb]
    \centering
    \begin{tabular}{p{3.5cm} p{5.2cm} p{5.2cm}}
        \cline{1-3}
        \textbf{Paper} & \textbf{Key insights} & \textbf{Applicability} \\
        \cline{1-3}
        \citet{zhai_scaling_2022} & ViTs follow power-law scaling but plateau at extreme compute levels, with benefits primarily seen in datasets >1B images. & Image classification, object detection, large-scale vision datasets. \\
        \cline{1-3}
        \citet{aghajanyan_scaling_2023} & Multimodal models experience competition at smaller scales but transition into synergy as model and token count grow, following a "competition barrier." & Multimodal learning, mixed-modal generative models, cross-domain AI. \\
        \cline{1-3}
        \citet{li_are_2024} & Scaling vision encoders in vision-language models (VLMs) does not always improve performance, reinforcing the importance of data quality over raw scaling. & Vision-language models, image-text alignment, multimodal scaling challenges. \\
        \cline{1-3}
    \end{tabular}
    \caption{Summary of key insights found in scaling laws paper for computer vision and multimodal domains.}
    \label{tab:scaling_laws2}
\end{table*}

%% file: tables/scaling_laws3.tex
\begin{table*}[!htb]
    \centering
    \scalebox{0.95}{
    \begin{tabular}{p{3.5cm} p{6cm} p{5.2cm}}
        \cline{1-3}
        \textbf{Paper} & \textbf{Key insights} & \textbf{Applicability} \\
        \cline{1-3}
        \multirow{4}{*}{\citet{hernandez_scaling_2021}} & Pre-training amplifies fine-tuning, particularly for small datasets, and benefits larger models even under data constraints. & Transfer learning, pre-training optimization, few-shot learning. \\
        \cdashline{1-3}
        \multirow{5}{*}{\citet{abnar_exploring_2021}} & Large-scale pre-training improves downstream performance, but effectiveness depends on upstream-downstream interactions, not task complexity. & Vision transfer learning, upstream-downstream performance interactions. \\
        \cdashline{1-3}
        \multirow{4}{*}{\citet{zhang_when_2024}} & Optimal fine-tuning strategy depends on dataset size: PEFT for small, LoRA for mid-scale, and full fine-tuning for large-scale datasets. & Fine-tuning strategies, parameter-efficient tuning, LoRA, full fine-tuning. \\
        \cdashline{1-3}
        \multirow{4}{*}{\citet{lin_selecting_2024}} & Fine-tuning follows a two-phase transition: slow early adaptation followed by power-law improvements, guiding compute-efficient model selection. & Compute-efficient fine-tuning, early stopping, model selection strategies. \\
        \cline{1-3}
    \end{tabular}
    }
    \caption{Key highlights from scaling of fine-tuned and domain-adapted models.}
    \label{tab:scaling_laws3}
\end{table*}

%% file: tables/scaling_laws4.tex
\begin{table*}[!htb]
    \centering
    \scalebox{0.9}{
    \begin{tabular}{p{3.5cm} p{6.3cm} p{5.9cm}}
        \cline{1-3}
        \textbf{Paper} & \textbf{Key insights} & \textbf{Applicability} \\
        \cline{1-3}
        \multirow{5}{*}{\citet{ brown_large_2024}} & Adaptive test-time compute strategies reduce computational costs by 4$\times$ while maintaining performance, enabling smaller models to compete with much larger ones. & Test-time compute efficiency, inference cost reduction, compute-limited environments. \\
        \cdashline{1-3}
        \multirow{4}{*}{\citet{wu_inference_2024}} & Advanced inference methods like REBASE tree search allow smaller models to match the performance of significantly larger ones. & High-efficiency inference, performance optimization for small models. \\
        \cdashline{1-3}
        \multirow{4}{*}{\citet{shao_scaling_2024}} & Increasing datastore size in retrieval-augmented models consistently improves performance under the same compute budget, without evident saturation. & Retrieval-augmented language models, knowledge-intensive tasks, compute-efficient architectures. \\
        \cdashline{1-3}
        \multirow{3}{*}{\citet{clark_unified_2022}} & Routing-based models show diminishing returns at larger scales, requiring optimal routing strategies for efficiency. & Routing-based models, MoEs, transformer scaling. \\
        \cdashline{1-3}
        \multirow{3}{*}{\citet{krajewski_scaling_2024}} & Fine-grained MoEs achieve up to 40$\times$ compute efficiency gains when expert granularity is optimized. & Mixture of Experts models, large-scale compute efficiency. \\
        \cdashline{1-3}
        \multirow{3}{*}{\citet{frantar_scaling_2023}} & Sparse model scaling enables predicting optimal sparsity levels for given compute budgets. & Sparse models, structured sparsity optimization, parameter reduction. \\
        \cline{1-3}
    \end{tabular}
    }
    \caption{Scaling laws of efficient models.}
    \label{tab:scaling_laws4}
\end{table*}

%% file: tables/scaling_laws5.tex
\begin{table*}[!htb]
    \centering
    \begin{tabular}{p{3.5cm} p{5.2cm} p{5.2cm}}
        \cline{1-3}
        \textbf{Paper} & \textbf{Key insights} & \textbf{Applicability} \\
        \cline{1-3}
        \citet{clark_unified_2022} & Routing-based models show diminishing returns at larger scales, requiring optimal routing strategies for efficiency. & Routing-based models, MoEs, transformer scaling. \\
        \cline{1-3}
        \citet{krajewski_scaling_2024} & Fine-grained MoEs achieve up to 40$\times$ compute efficiency gains when expert granularity is optimized. & Mixture of Experts models, large-scale compute efficiency. \\
        \cline{1-3}
        \citet{frantar_scaling_2023} & Sparse model scaling enables predicting optimal sparsity levels for given compute budgets. & Sparse models, structured sparsity optimization, parameter reduction. \\
        \cline{1-3}
    \end{tabular}
    \caption{Scaling laws for routing, sparsity, pruning, and quantization.}
    \label{tab:scaling_laws5}
\end{table*}

%% file: tables/scaling_laws6.tex
\begin{table*}[!htb]
    \centering
    \scalebox{0.95}{
    \begin{tabular}{p{3.5cm} p{5.2cm} p{5.9cm}}
        \cline{1-3}
        \textbf{Paper} & \textbf{Key insights} & \textbf{Applicability} \\
        \cline{1-3}
        \multirow{4}{*}{\citet{ye_data_2024}} & Predicts optimal data compositions before training, reducing compute costs by up to 27\% while maintaining performance. & Pre-training optimization, data efficiency improvements. \\
        \cdashline{1-3}
        \multirow{3}{*}{\citet{liu_regmix:_2024}} & REGMIX optimizes data mixtures using proxy models, achieving 90\% compute savings. & Compute-efficient training, automated data selection, large-scale models. \\
        \cdashline{1-3}
        \multirow{4}{*}{\citet{allen-zhu_physics_2024}} & Language models can store 2 bits of knowledge per parameter, with knowledge retention dependent on training exposure. & Knowledge encoding, model compression, retrieval-augmented models. \\
        \cline{1-3}
    \end{tabular}
    }
     \caption{Critical scaling laws for data mixing and knowledge capacity.}
     \label{tab:scaling_laws6}
\end{table*}

%% file: tables/database_desc1.tex
\begin{table*}[htbp]
  \centering
  
  \scalebox{0.5}{
    \begin{tabular}{p{30mm} p{30mm} p{50mm} p{50mm} p{80mm} rr}
    \cline{1-7}
    \textbf{Paper} & \textbf{Category} & \textbf{Task} & \textbf{Architecture} & \textbf{Datasets Used} & \textbf{Model Range} & \textbf{Data Range} \\
    \cline{1-7}
    \citet{kaplan_scaling_2020} & Pre-Training & Language Generation & Decoder-only Transformer & WebText2 & 0M - 1B & 22M - 23B \\
    \citet{hoffmann_training_2022} & Pre-Training  & Language Generation & Decoder-only Transformer & MassiveText,Github, C4 & 70M - 16B & 5B - 500B \\
    \citet{tay_scaling_2022} & Pre-Training , Transfer Learning & Language Generation & Switch, T5 Encoder-Decoder, Funnel, MoS, MLP-mixer, GLU, Lconv, Evolved, Dconv, Performer,Universal, ALBERT 
  & Pretraining: C4, Fine-Tuning: GLUE, SuperGLUE, SQuAD & 173M - 30B &  \\
    \citet{hu_minicpm:_2024} & Pre-Training & Language Generation & Decoder-only Transformer & Large mixture & 40M - 2B &  \\
    \citet{caballero_broken_2023} & Pre-Training & Downstream Image Recognition and Language Generation & ViT, Transformers, LSTM & Vision pretrained: JFT-300M, downstream : Birds200, Caltech101, CIFAR-100; Language : BigBench &       &  \\
    \citet{hernandez_scaling_2021} & Transfer Learning & Code Generation & Decoder-only Transformer & Pre-train: WebText2, CommonCrawl, English Wikipedia, Books; FineTune: Github repos &   &  \\
    \citet{abnar_exploring_2021} & Transfer Learning & Image Recognition & ViT, MLP-Mixers, ConvNets & Pre-train: JFT, ImageNet21K & 10M - 10B &  \\
    \citet{mikami_scaling_2021} & Transfer learning & Image Recognition & ConvNets &    Syntheic Data   &       &  \\
    \citet{zhang_when_2024} & Transfer Learning & Machine Translation and Language Generation & Decoder-only Transformer &  WMT14 English-German (En-De) and WMT19 English-Chinese (En-Zh), CNN/Daily-Mail, MLSUM  & 1B - 16B & 84B - 283B \\
    \citet{chen_scaling_2024} & Transfer learning & Language Generation & Decoder-only Transformer & Pre-Train: RedPajama v1, Validation: GitHub,ArXiv,Wikipedia, C4, RedPajama validation sets, ProofPile & 43M - 3B &  \\
    \citet{lin_selecting_2024} & Transfer learning & Language Generation & Decoder-only Transformer, Encoder-Decoder Transformer, Multilingual, MoE & Fine Tune: WMT19 English-Chinese (En-Zh), Gigaword, FLAN & 100M - 7B &  \\
    \citet{dettmers_case_2023} & Quantization Inference & Language Generation & Decoder-only Transformer &  The Pile, Lambada, PiQA, HellaSwag, Windogrande     & 19M - 176B &  \\
    \citet{cao_scaling_2024} & Quantization Inference & Language Generation & Decoder-only Transformer & WikiText2, SlimPajama, MMLU, Alpaca & 500M - 70B &  \\
    \citet{kumar_scaling_2024} & Quantization Pre-Training, Quantization Inference & Language Generation & Decoder-only Transformer & Dolma V1.7 & 30M - 220M & 1B - 26B \\
    \citet{chen_are_2024} & Inference & Language Generation & Decoder-only Transformer & MMLU Physics, TruthfulQA, GPQA, Averitec &       &  \\
    \citet{snell_scaling_2024} & Inference & Language Generation & Decoder-only Transformer & MATH &       &  \\
    \citet{brown_large_2024} & Inference & Language Generation & Decoder-only Transformer & GSM8K, MATH, MiniF2F-MATH, CodeContests, SWE-bench lite & 70M - 70B &  \\
    \citet{wu_inference_2024} & Inference & Language Generation & Decoder-only Transformer & MATH500, GSM8K & 410M - 34B &  \\
    \citet{sardana_beyond_2024} & Inference &    Language Generation   &  Decoder-only Transformer   & Jeopardy, MMLU, BIG bench, WikiData, ARC, COPA, PIQA, OpenBook QA, AGI Eval, GSM8k, etc  & 150M-6B   & 1.5B - 1.2T  \\
    \citet{clark_unified_2022} & Sparsity & Language Generation & Decoder-only Transformer, MoE &  MassiveText   & 0 - 200B & 0-130B  \\
    \citet{frantar_scaling_2023} & Sparsity & Language Generation, Image Recognition & Encoder-decoder, ViT & JFT-4B, C4 & 1M - 85M & 0 - 1B \\
    \citet{krajewski_scaling_2024} & Sparsity & Language generation & Decoder-only Transformer, MoE & C4   & 129M - 3B & 16B - 130B \\
    \citet{yun_toward_2024} & Sparsity  & Language generation & Decoder-only Transformer, MoE & Slim Pajama & 100M - 730M & 2B - 20B \\
    \citet{chen2024scaling} & Sparsity & Language Generation &  Decoder-only Transformer   &   SlimPajama    & 500M - 8B   & 0.5B \\
    \citet{busbridge2025distillation} & Distillation  & Language generation & Teacher-Student Decoder-only Transformer & C4 & 100M - 12B & 0 - 500B \\
    \citet{henighan_scaling_2020} & Multimodality & Generative Image Modeling, Video Modeling, Language Generation & Decoder-only Transformer &  FCC100M, and various modal datasets  & 0.1M-100B      & 100M \\
    \citet{zhai_scaling_2022} & Multimodality & Image Recognition & ViT & ImageNet-21K & 5M - 2B & 1M - 3B \\
    \citet{alabdulmohsin_revisiting_2022} & Multimodality & Image Recognition, Machine Translation & ViT, MLP Mixers, Encoder-decoder, Decoder-only Transformer, Transformer encoder-LSTM decoder & JFT-300M, ImageNet, Birds200, CIFAR100, Caltech101, Big-Bench  & 10M-1B     &  32M-494M\\
    \citet{aghajanyan_scaling_2023} & Multimodality & Multimodal Tasks & Decoder-only Transformers & OPT, Common Crawl, LibriSpeech , CommonVoice, VoxPopuli, Spotify Podcast, InCoder, SMILES from Zincand People’s Speech & 8M - 30B & 5B - 100B \\
    \citet{li_are_2024} & Multimodality & Multimodal tasks & ViT, Decoder-only Transformer & CC12M, LAION-400M & 7B - 13B & 1M - 10M \\
    \citet{jones_scaling_2021} & Multi-agent RL & Hex & AlphaZero with neural networks &       &       &  \\
    \citet{neumann_scaling_2023} & Multi-agent RL & Pentago, ConnectFour & AlphaZero with neural networks &       &       &  \\
    \citet{gao_scaling_2022} & RL    & Reward Model training with Best of n or RL & Decoder-only Transformers &       &       &  \\
    \citet{hilton_scaling_2023} & Single-agent RL & ProcGen Benchmark, 1v1 version of Dota2, toy MNIST & ConvNets, LSTM &       & 0M - 10M &  \\
    \citet{ye_data_2024} & Data Mixture & Language Generation & Decoder-only Transformer & RedPajama & 70M - 410M &  \\
    \citet{liu_regmix:_2024} & Data Mixture & Language Generation & Decoder-only Transformer & Pile &       &  \\
    \citet{kang_autoscale:_2024} & Data Mixture & Language Generation & Decoder-only Transformer , Encoder-only Transformer & RedPajama &       &  \\
    \citet{que_d-cpt_2024} & Data Mixture & Language Generation, Continual Pre-training & Decoder-only Transformer &  various mixture of Code, Math, Law, Chemistry, Music, Medical    &  0.5B-4B   & 0.1B-26B \\
    \citet{tao_scaling_2024} & Vocabulary & Language Generation & Decoder-only Transformer &  SlimPajama  & 33M - 3B & 0 - 500B \\
    \citet{circuits_nodate} & Sparse Autoencoder & Training Autoencoder & Decoder-only Transformer &       &       &  \\
    \citet{gao_scaling_2024} & Sparse Autoencoder & Find Interpretable Latents & Decoder-only Transformer &       &       &  \\
    \citet{shao_scaling_2024} & Retrieval & Language Generation & Decoder-only Transformer & language modelling:RedPajama, S2ORC,  Downstream : TriviaQA, NQ, MMLU, MedQA &       &  \\
    \citet{muennighoff_scaling_2023} & Pre-Training & Language Generation & Decoder-only transformer & C4 & 10M - 9B & 0 - 900B \\
    \citet{allen-zhu_physics_2024} & Knowledge Capacity & Language Generation & Decoder-only transformer &   bioD    &       &  \\
    \citet{ma_neural_2024} & Graph Supervised learning & Graph Classification Task & InfoGraph, GraphCL, JOAO, GraphMAE & reddit-threads , ogbg-molhiv,ogbg-molpcba &       &  \\
    \citet{diaz_scaling_2024} & Criticize &       &       &       &       &  \\
    \citet{sorscher_beyond_2023} & Criticize & Image Recognition & ConvNets, ViT & SVHN, CIFAR-10, and ImageNet &       &  \\
    \citet{bahri_explaining_2021} & Theoretical &       &       &       &       &  \\
    \citet{bordelon_dynamical_2024} & Theoretical &       &       &       &       &  \\
    \citet{hutter_learning_2021} & Theoretical &       &       &       &       &  \\
    \citet{lin2024scalinglawslinearregression} & Theoretical &       &       &       &       &  \\
    \citet{sharma_neural_2020} & Theoretical &       &       &       &       &  \\
    \citet{jin_cost_2023} & Downscaling &       &       &       &       &  \\
    \cline{1-7}
    \end{tabular}%
    }
    \caption{Details on task, architecture of models and training setup for each paper surveyed.}
  \label{tab:database1}%
\end{table*}%

%% file: tables/database_desc2.tex
\begin{table*}[htbp]
  \centering
  
  \scalebox{0.64}{
    \begin{tabular}{p{35mm} p{35mm} p{45mm} p{125mm}}
    \cline{1-4}
    Paper & Dependent variable & Scaling variable & Functional form \\
     \cline{1-4}
    \citet{kaplan_scaling_2020} & Pre-Training Loss & Model Parameters, Compute, Data, Training Steps& \(L(N,D) = \left[\left(\frac{N_c}{N}\right)^{\frac{\alpha_N}{\alpha_D}} + \frac{D_c}{D}\right]^{\alpha_D} \) \\
    \citet{hoffmann_training_2022} & Pre-Training Loss & Model Parameters, Data & \(L(N,D) = \frac{A}{N^\alpha} + \frac{B}{D^\beta} + E\) \\
    \citet{tay_scaling_2022} & Performance metric & Compute & \( P \propto  C^\alpha \) \\
    \citet{hu_minicpm:_2024} & Pre-Training Loss & Model Parameters, Data & \(L(P,D) = \frac{A}{N^\alpha} + \frac{B}{D^\beta} + E\) \\
    \citet{caballero_broken_2023} & Performance metric 
    & Model Parameters, Compute, Data, Input Size, Training Steps & \(y = a + \left(bx^{-c_0}\right)\prod_{i=1}^n\left(1 + \left(\frac{x}{d_i}\right)^{1/f_i}\right)^{-c_i*f_i}\) \\
    \citet{hernandez_scaling_2021} & Data Transferred & Model Parameters, Fine-tuning Data & \(D_t(D_f,N) = k(D_f)^\alpha(N)^\beta\) \\
    \citet{abnar_exploring_2021} & Downstream Error & Upstream Error & \(e_{DS} = k(e_{US})^a + c\) \\
    \citet{mikami_scaling_2021} & Downstream Error & Pre-training Data & \(e_{DS} = aD^{-\alpha} + c\) \\
    \citet{zhang_when_2024} & Downstream Loss & Fine-tuning Data, Data, Model Parameters, PET parameter & \(\hat{L}(X, D_f) = A * \frac{1}{X^\alpha} * \frac{1}{D_f^\beta} + E\) \\
    \citet{chen_scaling_2024} & Downstream performance & Pre-training Loss, Compute & \(L(C) = (\frac{C}{C_N})^\alpha\); \(P(L) = w_0 + w1 \cdot L\)  \\
    \citet{lin_selecting_2024} & Downstream Loss & Data, Fine-tuning Data & \(L(D) = \frac{B}{D_t + D^\beta} + E\) \\
    \citet{dettmers_case_2023} & Accurancy & Total Model Bits After Quantization &  \\
    \citet{cao_scaling_2024} & Total parameters & Quantization Ratio &  \\
    \citet{kumar_scaling_2024} & Loss & Data, Model Parameters, Training Precision, Post-train Precision & \(L(N, D, P_w, P_a, P_{\text{kv}}, P_{\text{post}}) =AN_{\text{eff}}^{-\alpha} + BD^{-\beta} + E + \delta_{PTQ}\) \\
    \citet{chen_are_2024} & Optimal LLM Calls & Fraction Of Easy And Difficult Queries &  \\
    \citet{brown_large_2024} & Coverage & Number Of Samples & \(\log(C) = ak^{-b}\) \\
    \citet{wu_inference_2024} & Optimal Compute   & Model Parameters & \(\log_{10}(C) = 1.19\log_{10}(N) + 2.03\) \\
    \citet{sardana_beyond_2024} & Pre-Training Loss & Model Parameters, Data & \(L(N,D) = \frac{A}{N^\alpha} + \frac{B}{D^\beta} + E\) \\
    \citet{clark_unified_2022} & Loss & Model Parameters, Number Of Experts , Data & \(\log(L(N,E)) = a\log N + b\log E + c\log N\cdot\log E + d\)  \\
    \citet{frantar_scaling_2023} & Loss & Sparsity, Model Parameters, Data & \(L=(a_S(1-S)^{b_S} + c_S) \cdot \left(\frac{1}{N}\right)^{b_N} + \left(\frac{a_D}{D}\right)^{b_D} + c\) \\
    \citet{krajewski_scaling_2024} & Loss & Granularity, Model Parameters, Data & \(\mathcal{L}(N,D,G) = c + \left(\frac{g}{G^\gamma} + a\right)\frac{1}{N^\alpha} + \frac{b}{D^\beta}\) \\
    \citet{yun_toward_2024} & Loss & Model Parameters, Number Of Experts , Data & \( \log L(N,D,E) \triangleq \log\left(\frac{A}{N^\alpha} + \frac{B}{E^\beta} + \frac{C}{D^\gamma} + F\right) + d\log N\log E \)\\
    \citet{chen2024scaling} &  Post-Training Loss  & Uncompressed Model Loss, pruned ratio, Model parameters before pruning, Post-training Data      & \(L(N_0, D, \rho, L_0) =  L_0 + \left( \frac{1}{\rho} \right)^\gamma  \left( \frac{1}{N_0} \right)^\delta  \left( \frac{N_C}{N_0^\alpha} + \frac{D_C}{D^\beta} + E \right)\) \\
    \citet{henighan_scaling_2020} & Loss & Model Parameters, Compute, Data & \(L(x) = Ax^{-\alpha} + B\) \\
    \citet{zhai_scaling_2022} & Downstream Error & Compute & \(E = aC^b +c\) \\
    \citet{alabdulmohsin_revisiting_2022} & Loss & Compute, Model Parameters, Data & \(\frac{L_x - L_\infty}{(L_0 - L_x)^\alpha} = \beta x^c\) \\
    \citet{aghajanyan_scaling_2023} & Loss & Model Parameters, Data & \(\mathcal{L}(N, D_i, D_j) = \left[\frac{\mathcal{L}(N, D_i) + \mathcal{L}(N, D_j)}{2}\right] - C_{i,j} + \frac{A_{i,j}}{N^{\alpha_{i,j}}} + \frac{B_{i,j}}{|D_i| + |D_j|^{\beta_{i,j}}}\) \\
    \citet{li_are_2024} & Loss & Model Parameters, Data &  \\
    \citet{jones_scaling_2021} & Elo & Compute, Board Size & \(Elo = \left( m_{\text{boardsize}}^{\text{plateau}} \cdot \text{boardsize} + c^{\text{plateau}}\right) \cdot clamp( m_{\text{boardsize}}^{\text{incline}} \cdot \text{boardsize} + m_{\text{flops}}^{\text{incline}} \cdot \log \text{flop} + c^{\text{incline}}, 0)\) \\
    \citet{neumann_scaling_2023} & Game Score  & Model Parameters, Compute & \(E_i = \frac{1}{1 + (X_j/X_i)^{\alpha_X}}\) \\
    \citet{gao_scaling_2022} & Gold Reward model scores & Root Of KL Between Initial Policy And Optimized Policy (d) & \(R(d) = d(\alpha - \beta\log d)\) \\
    \citet{hilton_scaling_2023} & Intrinsic performance 
    & Model Parameters, Environment Interactions 
    & \(I^{-\beta} = \left(\frac{N_c}{N}\right)^{\alpha_N} + \left(\frac{E_c}{E}\right)^{\alpha_E}\) \\
    \citet{ye_data_2024} & Loss on domain i & Proportion Of Training Domains & \(L_i(r_{1...M}) = c_i + k_i \exp\left(\sum_{j=1}^M t_{ij}r_j\right)\) \\
    \citet{que_d-cpt_2024} & Validation loss & Model Parameters, Data, Mixture Ratio & \(L(N, D, r) = E + \frac{A}{N^\alpha} + \frac{B \cdot r^\eta}{D^\beta} + \frac{C}{(r + \epsilon)^\gamma}\) \\
    \citet{tao_scaling_2024} & Unigram-Normalised loss & Non-vocabulary Parameter, Vocabulary Parameters, Data & \(\mathcal{L}_u = -E + \frac{A_1}{N_{\text{nv}}^{\alpha_1}} + \frac{A_2}{N_{\text{v}}^{\alpha_2}} + \frac{B}{D^\beta}\) \\
    \citet{circuits_nodate} & Reconstruction error & Compute, Number Of Latents &  \\
    \citet{gao_scaling_2024} & Reconstruction loss & Number Of Latents, Sparsity Level & \(L(n,k) = \exp(\alpha + \beta_k \log(k) + \beta_n \log(n) + \gamma \log(k)\log(n)) + \exp(\zeta + \eta \log(k))\) \\
    \citet{shao_scaling_2024} &  Downstream Accuracy   & Datastore , Model Parameters, Data, Compute    &  \\
    \citet{muennighoff_scaling_2023} & Loss & Data, Model Parameters, Epochs & \(L(N,D) = \frac{A}{N'^\alpha} + \frac{B}{D'^\beta} + E\) \\
    \citet{busbridge2025distillation} & Student Loss & Teacher Loss, Student Parameters, Distillation Tokens  & \(L_S(N_S,D_S,L_T) = L_T + \frac{1}{L_T^{c_0}} \left(1 + \left(\frac{L_T}{L_{S,d_1}}\right)^{1/f_1}\right)^{-c_1/f_1} \left(\frac{A}{N_S^{\alpha'}} + \frac{B}{D_S^{\beta'}}\right)^{\gamma'}\) \\
     \cline{1-4}
    \end{tabular}%
    }
    \caption{Scaling law forms proposed in different papers we surveyed.}
  \label{tab:database2}%
\end{table*}%

%% file: tables/github_link.tex
\begin{table*}[htbp]
    \centering
    \scalebox{0.9}{
    \begin{tabular}{lccc}
        \cline{1-4}
        \textbf{Paper} & \textbf{Training code} & \textbf{Analysis code} & \textbf{Github link} \\ \cline{1-4}
        \citet{kaplan_scaling_2020} & N & N & ~ \\ 
        \citet{hoffmann_training_2022} & N & N & ~ \\ 
        \citet{hoffmann_training_2022} & N & N & ~ \\ 
        \citet{hu_minicpm:_2024} & Y & N & \href{https://github.com/OpenBMB/MiniCPM}{Link} \\ 
        \citet{caballero_broken_2023} & N & Y & \href{https://github.com/ethancaballero/broken_neural_scaling_laws}{Link} \\ 
        \citet{hernandez_scaling_2021} & N & N & ~ \\ 
        \citet{abnar_exploring_2021} & N & N & ~ \\ 
        \citet{mikami_scaling_2021} & N & Y & \href{https://github.com/pfnet-research/cg-transfer}{Link} \\ 
        \citet{zhang_when_2024} & N & N & ~ \\ 
        \citet{chen_scaling_2024} & N & N & ~ \\ 
        \citet{lin_selecting_2024} & N & Y & \href{https://github.com/linhaowei1/Fine-tuning-Scaling-Law/blob/main/benchmark/flan.csv}{Link}  \\ 
        \citet{dettmers_case_2023} & N & N & ~ \\ 
        \citet{cao_scaling_2024} & N & N & ~ \\ 
        \citet{kumar_scaling_2024} & N & N & ~ \\ 
        \citet{chen_are_2024} & Y & Y & \href{https://github.com/lchen001/CompoundAIScalingLaws}{Link}  \\ 
        \citet{snell_scaling_2024}& N & N & ~ \\ 
        \citet{brown_large_2024} & Y & Y & \href{https://github.com/ScalingIntelligence/large\_language\_monkeys/tree/main}{Link}  \\ 
        \citet{wu_inference_2024} & Y & N & \href{https://github.com/thu-wyz/inference_scaling}{Link}  \\ 
        \citet{sardana_beyond_2024} & N & N & ~ \\ 
        \citet{clark_unified_2022} & N & Y & \href{https://github.com/google-deepmind/scaling_laws_for_routing}{Link}  \\ 
        \citet{frantar_scaling_2023} & N & N & ~ \\ 
        \citet{krajewski_scaling_2024} & Y & Y & \href{https://github.com/llm-random/llm-random}{Link}  \\ 
        \citet{yun_toward_2024} & N & N & ~ \\ 
        \citet{chen2024scaling} & N & N & ~ \\ 
        \citet{henighan_scaling_2020} & N & N & ~ \\ 
        \citet{zhai_scaling_2022} & Y & N & \href{https://github.com/google-research/vision_transformer}{Link}  \\ 
        \citet{alabdulmohsin_revisiting_2022} & N & Y & \href{https://github.com/google-research/google-research/tree/master/revisiting_neural_scaling_laws}{Link}  \\ 
        \citet{aghajanyan_scaling_2023} & N & N & ~ \\ 
        \citet{li_are_2024} & N & N & ~ \\ 
        \citet{jones_scaling_2021} & Y & Y & \href{https://github.com/andyljones/boardlaw}{Link}  \\ 
        \citet{neumann_scaling_2023} & Y & Y & \href{https://github.com/OrenNeumann/AlphaZero-scaling-laws}{Link}  \\ 
        \citet{gao_scaling_2022} & N & N & ~ \\ 
        \citet{hilton_scaling_2023} & N & N & ~ \\ 
        \citet{ye_data_2024} & Y & Y & \href{https://github.com/yegcjs/mixinglaws}{Link}  \\ 
        \citet{liu_regmix:_2024} & Y & Y & \href{https://github.com/sail-sg/regmix}{Link}  \\ 
         \citet{kang_autoscale:_2024} & Y & Y & \href{https://github.com/feiyang-k/AutoScale }{Link} \\ 
        \citet{que_d-cpt_2024} & N & N & ~ \\ 
        \citet{tao_scaling_2024} & Y & Y & \href{https://github.com/sail-sg/scaling-with-vocab}{Link}  \\ 
        \citet{circuits_nodate} & N & N & ~ \\ 
        \citet{gao_scaling_2024} & Y & Y & \href{https://github.com/openai/sparse_autoencoder}{Link}  \\ 
        \citet{shao_scaling_2024} & Y & Y & \href{https://github.com/RulinShao/retrieval-scaling}{Link}  \\ 
        \citet{muennighoff_scaling_2023} & Y & Y & \href{https://github.com/huggingface/datablations}{Link}  \\ 
        \citet{allen-zhu_physics_2024} & N & N & ~ \\ 
        \citet{ma_neural_2024} & Y & N & \href{https://github.com/HaitaoMao/Graph-Neural-Scaling-Law}{Link}  \\ 
        \citet{sorscher_beyond_2023} & N & Y & \href{https://github.com/rgeirhos/dataset-pruning-metrics}{Link}  \\
        \cline{1-4}
    \end{tabular}
    }
    \caption{Reproducibility of different neural scaling law papers. Reproducibility status of 45 papers surveyed: 22 (48.9\%) provided repositories; 29 (64.4\%) did not share training code.}
    \label{tab:reproducibility}
\end{table*}